%% file: main.tex
\documentclass[pdflatex,sn-basic,Numbered,iicol]{sn-jnl}
\usepackage[T1]{fontenc}
\usepackage[utf8]{inputenc}

\usepackage{anyfontsize}

\usepackage{packages/main}
\usepackage{packages/macros}
\usepackage{packages/theorems}

\loadglsentries{glossaries/abbreviations}
\loadglsentries[symbols]{glossaries/symbols}
\loadglsentries[logic]{glossaries/logic}
\loadglsentries[keyterms]{glossaries/keyterms}

\begin{document}

\input{src/header}
\input{src/abstract}\glsresetall
\maketitle

\input{src/sections}

\bibliography{main}



\end{document}

%% file: glossaries/abbreviations.tex
\setabbreviationstyle[tool]{long-short-sc}
\newabbreviation[category=tool]{a:strem}{Strem}{Spatio-Temporal Regular Expression Matcher}
\newabbreviation[category=tool]{a:scenic}{Scenic}{Scenario Description Language}
\newabbreviation{a:spre}{SpRE}{Spatial Regular Expression}
\newabbreviation{a:re}{RE}{Regular Expression}
\newabbreviation{a:dfa}{DFA}{Deterministic Finite Automata}
\newabbreviation{a:ltl}{LTL}{Linear Temporal Logic}
\newabbreviation{a:cv}{CV}{Computer Vision}
\newabbreviation{a:ads}{ADS}{Autonomous Driving System}
\newabbreviation{a:av}{AV}{Autonomous Vehicle}
\newabbreviation{a:ros}{ROS}{Robot Operating System}
\newabbreviation{a:carla}{CARLA}{Car Learning to Act}
\newabbreviation{a:ml}{ML}{Machine Learning}
\newabbreviation{a:ad}{AD}{Autonomous Driving}
\newabbreviation{a:veql}{VEQL}{Video Event Query Language}
\newabbreviation{a:aia}{AIA}{Allen Interval Algebra}
\newabbreviation{a:tqtl}{TQTL}{Timed Quality Temporal Logic}
\newabbreviation{a:stpl}{STPL}{Spatio-Temporal Perception Logic}
\newabbreviation{a:smt}{SMT}{Satisfiability Modulo Theory}
\newabbreviation{a:bm}{BM}{Boyer-Moore}
\newabbreviation{a:kmp}{KMP}{Knuth-Morris-Pratt}
\newabbreviation{a:cctv}{CCTV}{Closed-Circuit Television}
\newabbreviation{a:cli}{CLI}{Command Line Interface}
\newabbreviation{a:tcp}{TCP}{Transmission Control Protocol}
\newabbreviation{a:s-ast}{s-AST}{symbolic-Abstract Syntax Tree}
\newabbreviation{a:ir}{IR}{Intermediate Representation}
\newabbreviation{a:lidar}{LiDAR}{Light Detecting and Radar}
\newabbreviation{a:dnn}{DNN}{Deep Neural Network}
\newabbreviation{a:dfs}{DFS}{Depth-First Search}
\newabbreviation{a:nfa}{NFA}{Nondeterministic Finite Automata}
\newabbreviation{a:gps}{GPS}{Global Positioning System}
\newabbreviation{a:imu}{IMU}{Inertial Measurement Unit}
\newabbreviation{a:iou}{IoU}{Intersection over Union}
\newabbreviation{a:aabb}{AABB}{Axis Aligned Bounding Box}
\newabbreviation{a:wff}{WFF}{Well-Formed Formula}

%% file: glossaries/symbols.tex
\glsxtrnewsymbol[description={Alphabet}]{s:alphabet}{\ensuremath{\Sigma}}
\glsxtrnewsymbol[description={Boolean Set}]{s:booleans}{\ensuremath{\mathbb{B}}}
\glsxtrnewsymbol[description={True}]{s:true}{\ensuremath{\top}}
\glsxtrnewsymbol[description={False}]{s:false}{\ensuremath{\bot}}
\glsxtrnewsymbol[description={Integer Set}]{s:integers}{\ensuremath{\mathbb{Z}}}
\glsxtrnewsymbol[description={Natural Number Set}]{s:naturals}{\ensuremath{\mathbb{N}}}
\glsxtrnewsymbol[description={Real Number Set}]{s:reals}{\ensuremath{\mathbb{R}}}
\glsxtrnewsymbol[description={Powerset}]{s:powerset}{\ensuremath{\mathcal{P}}}
\glsxtrnewsymbol[description={Perception Stream}]{s:datastream}{\ensuremath{\mathcal{S}}}
\glsxtrnewsymbol[description={Frame}]{s:frame}{\ensuremath{\mathcal{F}}}
\glsxtrnewsymbol[description={Logical Not}]{s:not}{\ensuremath{\neg}}
\glsxtrnewsymbol[description={Logical And}]{s:and}{\ensuremath{\land}}
\glsxtrnewsymbol[description={Logical Or}]{s:or}{\ensuremath{\lor}}
\glsxtrnewsymbol[description={\ensuremath{\mathcal{S}4} Logic}]{s:s4}{\ensuremath{\mathcal{S}4}}
\glsxtrnewsymbol[description={\ensuremath{\mathcal{S}4_{u}}} Logic]{s:s4u}{\ensuremath{\glsentryname{s:s4}_{u}}}
\glsxtrnewsymbol[description={\ensuremath{\mathcal{S}4^{+}}} Logic]{s:s4+}{\ensuremath{\glsentryname{s:s4}^{+}}}
\glsxtrnewsymbol[description={\ensuremath{\mathcal{S}4^{+}_{u}} Logic}]{s:s4u+}{\ensuremath{\glsentryname{s:s4u}^{+}}}
\glsxtrnewsymbol[description={Spatial Term}]{s:s4+form}{\ensuremath{\tau}}
\glsxtrnewsymbol[description={Spatial Formula}]{s:s4u+form}{\ensuremath{\phi}}
\glsxtrnewsymbol[description={SpRE Pattern}]{s:spreform}{\ensuremath{P}}
\glsxtrnewsymbol[description={State}]{s:states}{\ensuremath{\pi}}
\glsxtrnewsymbol[description={Annotations}]{s:annot}{\ensuremath{\alpha}}
\glsxtrnewsymbol[description={Intersection}]{s:inter}{\ensuremath{\sqcap}}
\glsxtrnewsymbol[description={Union}]{s:union}{\ensuremath{\sqcup}}
\glsxtrnewsymbol[description={Interior}]{s:interior}{\ensuremath{\mathbf{I}}}
\glsxtrnewsymbol[description={Closure}]{s:closure}{\ensuremath{\mathbf{C}}}
\glsxtrnewsymbol[description={RE $\times$ $\mathcal{S}4_{u}$}]{s:rexs4u}{\ensuremath{\textnormal{RE}\times\glsentryname{s:s4u}}}
\glsxtrnewsymbol[description={Non-Empty}]{s:isnonempty}{\ensuremath{\scalebox{0.65}{\boxed{\exists}}}}
\glsxtrnewsymbol[description={Subset}]{s:issubset}{\ensuremath{\sqsubseteq}}
\glsxtrnewsymbol[description={Alternation}]{s:alt}{\ensuremath{\rule[0em]{0.4pt}{0.55em}}}
\glsxtrnewsymbol[description={Concatenation}]{s:concat}{\ensuremath{\cdot}}
\glsxtrnewsymbol[description={Bounding Box, Function}]{s:bbox}{\ensuremath{\mathfrak{B}}}
\glsxtrnewsymbol[description={Attributes Function}]{s:attrs}{\ensuremath{\mathfrak{A}}}
\glsxtrnewsymbol[description={Set-Theoretic Intersection}]{s:sinter}{\ensuremath{\cap}}
\glsxtrnewsymbol[description={Set-Theoretic Union}]{s:sunion}{\ensuremath{\cup}}
\glsxtrnewsymbol[description={Satisfies}]{s:satis}{\ensuremath{\vDash}}
\glsxtrnewsymbol[description={Language}]{s:langof}{\ensuremath{\mathcal{L}}}
\glsxtrnewsymbol[description={Symbol}]{s:symbol}{\ensuremath{\sigma}}
\glsxtrnewsymbol[description={Query}]{s:query}{\ensuremath{Q}}
\glsxtrnewsymbol[description={Kleene-star}]{s:kleene}{\ensuremath{^{*}}}
\glsxtrnewsymbol[description={Infinity}]{s:infty}{\ensuremath{\infty}}
\glsxtrnewsymbol[description={Subsequence Operator}]{s:subseq}{\ensuremath{\preceq}}
\glsxtrnewsymbol[description={Set-Theoretic Complement}]{s:complement}{\ensuremath{\bar{\phantom{x}}}}
\glsxtrnewsymbol[description={Object Set}]{s:objects}{\ensuremath{\mathcal{O}}}
\glsxtrnewsymbol[description={Subset-Equal}]{s:subseteq}{\ensuremath{\subseteq}}
\glsxtrnewsymbol[description={Not Equal}]{s:neq}{\ensuremath{\neq}}
\glsxtrnewsymbol[description={Empty Set}]{s:emptyset}{\ensuremath{\emptyset}}
\glsxtrnewsymbol[description={Timestamp Threshold}]{s:tsthresh}{\ensuremath{\epsilon_{t}}}
\glsxtrnewsymbol[description={Sensor Channels}]{s:channels}{\ensuremath{\mathcal{C}}}
\glsxtrnewsymbol[description={Sensor Channel}]{s:channel}{\ensuremath{c}}
\glsxtrnewsymbol[description={PTL $\times$ $\mathcal{S}4_{u}$}]{s:ptlxs4u}{\ensuremath{\textnormal{PTL}\times\glsentryname{s:s4u}}}
\glsxtrnewsymbol[description={Object}]{s:object}{\ensuremath{o}}
\glsxtrnewsymbol[description={Image Space}]{s:imagespace}{\ensuremath{\mathfrak{I}}}
\glsxtrnewsymbol[description={Workspace}]{s:workspace}{\ensuremath{X}}
\glsxtrnewsymbol[description={Reason Space}]{s:reasonspace}{\ensuremath{W}}
\glsxtrnewsymbol[description={Valuation}]{s:valuation}{\ensuremath{\llbracket\: \rrbracket}}
\glsxtrnewsymbol[description={Symbolic Formula}]{s:syform}{\ensuremath{\varphi}}
\glsxtrnewsymbol[description={Symbolic Symbol}]{s:ssymbol}{\ensuremath{\sigma_{s}}}
\glsxtrnewsymbol[description={Word}]{s:word}{\ensuremath{w}}
\glsxtrnewsymbol[description={Pattern}]{s:pattern}{\ensuremath{p}}
\glsxtrnewsymbol[description={Horizon}]{s:horizon}{\ensuremath{\mathcal{H}}}
\glsxtrnewsymbol[description={Symbol Alphabet}]{s:s-alphabet}{\ensuremath{\glsfmttext{s:alphabet}_{\phi}}}
\glsxtrnewsymbol[description={Model}]{s:model}{\ensuremath{M}}
\glsxtrnewsymbol[description={Spatial Formulas}]{s:s-formulas}{\ensuremath{\Phi}}
\glsxtrnewsymbol[description={State}]{s:state}{\ensuremath{q}}
\glsxtrnewsymbol[description={Transition Function}]{s:transfn}{\ensuremath{\delta}}
\glsxtrnewsymbol[description={Accepting States}]{s:accepting}{\ensuremath{F}}
\glsxtrnewsymbol[description={Environment}]{s:environment}{\ensuremath{\mathcal{E}}}
\glsxtrnewsymbol[description={Attributes Map}]{s:attributes}{\ensuremath{\mathcal{A}}}
\glsxtrnewsymbol[description={Attribute Keys}]{a:keys}{\ensuremath{\mathcal{K}}}
\glsxtrnewsymbol[description={Deterministic Finite Automata}]{s:dfa}{\ensuremath{D}}
\glsxtrnewsymbol[description={Order of}]{s:bigo}{\ensuremath{\mathcal{O}}}

\glsxtrnewsymbol[description={\ensuremath{\mathcal{S}4_{m}}} Logic]{s:s4m}{\ensuremath{\glsentryname{s:s4}_{m}}}
\glsxtrnewsymbol[description={\ensuremath{\mathcal{S}4^{+}_{m}} Logic}]{s:s4m+}{\ensuremath{\glsentryname{s:s4m}^{+}}}
\glsxtrnewsymbol[description={Spatial Expression}]{s:s4m+form}{\ensuremath{\psi}}

\glsxtrnewsymbol[description={Variables}]{s:vars}{\ensuremath{V}}
\glsxtrnewsymbol[description={Object Function}]{s:objfn}{\ensuremath{\mathfrak{R}}}
\glsxtrnewsymbol[description={Lookup Table}]{s:ltable}{\ensuremath{\mathfrak{T}}}

%% file: glossaries/logic.tex
\newglossary*{logic}{Logic}

\newglossaryentry{l:and}{%
    name={\ensuremath{\land}},
    description={logical and},
    type={logic}
}

\newglossaryentry{l:or}{%
    name={\ensuremath{\lor}},
    description={logical or},
    type={logic}
}

\newglossaryentry{l:not}{%
    name={\ensuremath{\neg}},
    description={logical not},
    type={logic}
}

%% file: glossaries/keyterms.tex
\newglossary*{keyterms}{Keyterms}

\newglossaryentry{k:range-op}{
  first={\textit{range}},
  name={range},
  description={},
  type={keyterms},
  category={keyterms}
}

\newglossaryentry{k:axis-aligned}{
  first={\textit{axis aligned}},
  name={axis aligned},
  description={},
  type={keyterms},
  category={keyterms}
}

\newglossaryentry{k:object-tracking}{
  first={\textit{object tracking}},
  name={object tracking},
  description={},
  type={keyterms},
  category={keyterms}
}

\newglossaryentry{k:key-frame}{
  first={\textit{key frame}},
  name={key frame},
  description={},
  type={keyterms},
  category={keyterms}
}

\newglossaryentry{k:reverse}{
  first={\textit{reverse}},
  name={reverse},
  description={},
  type={keyterms},
  category={keyterms}
}

\newglossaryentry{k:forward}{
  first={\textit{forward}},
  name={forward},
  description={},
  type={keyterms},
  category={keyterms}
}

\newglossaryentry{k:ego}{
  first={\textit{ego}},
  name={ego},
  description={},
  type={keyterms},
  category={keyterms}
}

\newglossaryentry{k:topics}{
  first={\textit{topics}},
  name={topics},
  description={},
  type={keyterms},
  category={keyterms}
}

\newglossaryentry{k:wall-clock}{
  first={\textit{wall clock}},
  name={wall clock},
  description={The \glsfmttext{a:strem} monitor.},
  type={keyterms},
  category={keyterms}
}

\newglossaryentry{k:dataset}{
  first={\textit{Woven Planet Perception}},
  name={Woven Planet Perception},
  description={The \glsfmttext{a:strem} monitor.},
  type={keyterms},
  category={keyterms}
}

\newglossaryentry{k:frontend}{
    first={\textit{frontend}},
    name={frontend},
    description={The frontend partition.},
    type={keyterms},
    category={keyterms}
}

\newglossaryentry{k:backend}{
    first={\textit{backend}},
    name={backend},
    description={The backend partition.},
    type={keyterms},
    category={keyterms}
}

\newglossaryentry{k:language-recognition}{
    first={\textit{language recognition}},
    name={language recognition},
    description={The language recognition problem.},
    type={keyterms},
    category={keyterms}
}

\newglossaryentry{k:matcher}{
    first={\added{\textit{Matcher}}},
    name={\replaced{Matcher}{\textit{Matcher}}},
    description={The \glsfmttext{a:strem} matcher.},
    type={keyterms},
    category={keyterms}
}

\newglossaryentry{k:monitor}{
    first={\added{\textit{Monitor}}},
    name={\replaced{Monitor}{\textit{Monitor}}},
    description={The \glsfmttext{a:strem} monitor.},
    type={keyterms},
    category={keyterms}
}

\newglossaryentry{k:compiler}{
    first={\added{\textit{Compiler}}},
    name={\replaced{Compiler}{\textit{Monitor}}},
    description={The \glsfmttext{a:strem} compiler.},
    type={keyterms},
    category={keyterms}
}

\newglossaryentry{k:rust}{
    first={\textit{Rust}},
    name={Rust},
    description={The programming language, Rust.},
    type={keyterms},
    category={keyterms}
}

\newglossaryentry{k:monitoring}{
    first={\textit{monitoring}},
    name={monitoring},
    description={The monitoring problem.},
    type={keyterms},
    category={keyterms}
}

\newglossaryentry{k:pattern-matching}{
    first={\textit{pattern matching}},
    name={pattern matching},
    description={The pattern matching problem},
    type={keyterms},
    category={keyterms}
}

\newglossaryentry{k:offline}{
    first={\textit{offline}},
    name={offline},
    description={All inputs are virtually available.},
    type={keyterms},
    category={keyterms}
}

\newglossaryentry{k:online}{
    first={\textit{online}},
    name={online},
    description={A virtually infinite series of inputs.},
    type={keyterms},
    category={keyterms}
}

\newglossaryentry{k:grep}{
    first={\textit{grep}},
    name={\texttt{grep}},
    description={A popular regular expression matcher and printer.},
    type={keyterms},
    category={keyterms}
}

\newglossaryentry{k:egrep}{
    first={\textit{egrep}},
    name={egrep},
    description={An extended version of grep.},
    type={keyterms},
    category={keyterms}
}

\newglossaryentry{k:charclass}{
    first={\textit{character class}},
    name={character class},
    description={A set of characters.},
    type={keyterms},
    category={keyterms},
    plural={character classes}
}

\newglossaryentry{k:brexpr}{
    first={\textit{bracket expression}},
    name={bracket expression},
    description={Another term for POSIX character class.},
    type={keyterms},
    category={keyterms}
}

\newglossaryentry{k:top-down}{
    first={\textit{top-down}},
    name={top-down},
    description={A perspective from which the camera is facing perpendicular to the x-y plane.},
    type={keyterms},
    category={keyterms}
}

\newglossaryentry{k:sensor-fusion}{
    first={\textit{sensor fusion}},
    name={sensor fusion},
    description={The procedure of utilizing multiple sensors to produce data.},
    type={keyterms},
    category={keyterms}
}

\newglossaryentry{k:spatial-term}{
    first={\textit{spatial term}},
    name={spatial term},
    description={A set-based formula.},
    type={keyterms},
    category={keyterms}
}

\newglossaryentry{k:metric-expression}{
    first={\textit{metric expression}},
    name={metric expression},
    description={A metric-based formula.},
    type={keyterms},
    category={keyterms}
}

\newglossaryentry{k:spatial-formula}{
    first={\textit{spatial formula}},
    name={spatial formula},
    description={A boolean-based formula.},
    type={keyterms},
    category={keyterms}
}

\newglossaryentry{k:query}{
    first={\textit{query}},
    name={query},
    description={A regular expression-based query (i.e., a SpRE).},
    type={keyterms},
    category={keyterms}
}

\newglossaryentry{k:verification}{
  first={\textit{verification}},
  name={verification},
  description={},
  type={keyterms},
  category={keyterms}
}

\newglossaryentry{k:validation}{
  first={\textit{validation}},
  name={validation},
  description={},
  type={keyterms},
  category={keyterms}
}

%% file: src/header.tex
\title{\replaced{Querying Perception Streams with \glsfmtlongpl{a:spre}}{Pattern Matching for Perception Streams}}

\author*[1]{\fnm{Jacob} \sur{Anderson}}\email{jacob.anderson@toyota.com}
\author[1]{\fnm{Georgios} \sur{Fainekos}}\email{georgios.fainekos@toyota.com}
\author[1]{\fnm{Bardh} \sur{Hoxha}}\email{bardh.hoxha@toyota.com}
\author[1]{\fnm{Hideki} \sur{Okamoto}}\email{hideki.okamoto@toyota.com}
\author[1]{\fnm{Danil} \sur{Prokhorov}}\email{danil.prokhorov@toyota.com}

\affil*[1]{\orgdiv{Future Research Department}, \orgname{Toyota Motor North America, Research \& Development}, \city{Ann Arbor}, \state{MI}, \country{USA}}


%% file: src/abstract.tex
\abstract{
  \added[id=gf]{Perception in fields like robotics, manufacturing, and data analysis generates large volumes of temporal and spatial data to effectively capture their environments. However, sorting through this data for specific scenarios is a meticulous and error-prone process, often dependent on the application, and lacks generality and reproducibility.}
  In this work, we introduce \glspl{a:spre} as a nov\-el querying language for pattern matching over perception streams containing spatial and temporal data derived from multi-modal dynamic environments.
  To highlight the capabilities of \glspl{a:spre}, we developed the \glsxtrshort{a:strem} tool as both an \gls{k:offline} and \gls{k:online} pattern matching framework for perception data.
  We demonstrate the offline capabilities of  \glsxtrshort{a:strem} through a case study on a publicly available \glsxtrshort{a:av} dataset (Woven Planet Perception) and its online capabilities through a case study integrating \glsxtrshort{a:strem} in  \glsxtrshort{a:ros} with the  \glsxtrshort{a:carla} simulator.
  We also conduct performance benchmark experiments on various \gls{a:spre} queries.
  Using our matching framework, we are able to find over 20,000 matches within 296 ms making \glsxtrshort{a:strem} applicable in runtime monitoring applications.
}

\keywords{Pattern matching, Regular expressions, Spatial logic, Computer vision, Runtime monitoring}


%% file: src/sections.tex
\input{src/sections/introduction}
\input{src/sections/preliminaries}
\input{src/sections/spre}
\input{src/sections/matching}
\input{src/sections/experiments}
\input{src/sections/literature}
\input{src/sections/conclusion}


%% file: src/sections/introduction.tex
\section{Introduction}
\label{sec:intro}

Perception systems are utilized across a wide range of applications---from \added{safety-critical systems such as} \glspl{a:av} \cite{janai2020computer,zhang2023perception,meng2023configuration}, to \added{informational systems such as} sports media analysis \cite{thomas2017computer,xu2016census}, to \added{security-based systems such as} \glspl{a:cctv} \cite{turtiainen2020towards}, and more \cite{fang2020computer,ward2021computer,kapach2012computer,lu2023carom}.
\added{To effectively perceive the environment,} \replaced{these}{These} systems may be composed of various sensor and \gls{k:sensor-fusion} technologies such as \glspl{a:lidar}, radars, cameras, etc. to support complex \gls{a:cv} tasks in both the \gls{k:offline} and \gls{k:online} domain that generate and require a significant amount of data to effectively operate \cite{bai2023cyber}.
To improve upon these perception systems and further \gls{a:ml} activities, large datasets are released for \gls{a:cv} tasks in hopes of providing more exposure to these systems before deployment \cite{lin2014microsoft,everingham2010pascal}.
These perception-based datasets are further extended to \gls{a:ads} applications where the perception system consists of a suite of sensors \added{and \glspl{k:sensor-fusion} in order to accurately capture and predict the highly dynamic scenarios encountered}.
Examples of such datasets include the popular Waymo Open \cite{sun2020scalability}, \replaced{Woven Planet Perception (Lyft) Level 5}{Woven Planet (``L5'') Perception} \cite{kesten2019woven}, and NuScenes \cite{caesar2020nuscenes} along with \replaced{many}{several} others \cite{pitropov2021canadian,yu2020bdd100k,xiao2021pandaset}.
Therefore, as these perception systems become more comprehensive\added{ and used}, methods and tools that enable querying over \replaced{the perception data generated}{such stream data} for specific scenarios in \textit{testing}, \textit{training}, and \gls{k:monitoring} become increasingly important.

For a given perception stream, however, filtering and searching for scenarios of interest is not well-supported nor a ubiquitous process as the size of data, selected schema, and present sensor suite varies \added{greatly}.
Within the \gls{k:offline} setting, perception streams produced by \gls{a:av} companies provide minimal frameworks to interface and filter data according to a pre-defined schema \added{without additional work to reason efficiently over the underlying data}.
As for the \gls{k:online} setting, perception systems streaming data in real-time are not traditionally responsible for identifying \added{particular} scenarios \added{of interest (e.g., for perception in \glspl{a:av}, identify scenarios where a bicyclist is near the ego car for ten seconds)}.
Therefore, this work aims to address the problem of querying complex and dynamic perception streams comprised of spatially- and temporally-aligned data \gls{k:offline} and \gls{k:online}.

In the work presented in this paper, we introduce \glspl{a:spre}: a novel querying language for efficient and flexible matching of perception streams. 
\glspl{a:spre} combine \glspl{a:re} \cite{aho1980pattern} with the modal logic of topology \gls{s:s4u} \cite{kontchakov2007spatial}.
The language is designed with ease-of-use in mind by following syntactic similiarities of classic \gls{a:re} tools such as \gls{k:grep} and \gls{k:egrep} \cite{friedl2006mastering} while enabling reasoning over topological relations. 
The querying language has been implemented in the \gls{a:strem} tool to support \gls{k:offline} and \gls{k:online} searching capabilities.
The \added{spatial-based} \gls{a:spre} queries can be efficiently solved due to the reduction of the pattern matching problem to the one for \replaced{\glspl{a:re}}{REs}, which in turn allows us to utilize well-established \added{algorithms and} libraries \cite{gallant2022regex} for fast \replaced{searching}{processing}.
Furthermore, the \gls{a:strem} tool's modular design and compatibility \replaced{as}{with} \deleted{Linux, Bash, and}\replaced{a}{any} \gls{a:cli} \replaced{tool}{tools} \replaced{makes}{make} the system more versatile and \replaced{applicable}{extendable} for use in \replaced{\gls{k:verification}}{\textit{verification}} and \replaced{\gls{k:validation}}{\textit{validation}} pipelines and frameworks.
Our formulation of the \gls{a:spre} language (\cref{sec:spre}) is also general enough to utilize other branches of logic without major restructuring.

\subsection{Summary of Contributions}
In this paper, we extend our prior work \cite{anderson2023pattern} as follows:
\begin{itemize}
  \item \added[id=gf]{We add metric expressions (Def. \ref{def:spre:syntax:2}) to the \gls{a:spre} language to support numerical expressions over functions with set-based domains.}
  \item \added[id=gf]{We introduce a \gls{k:top-down} perception environment as an additional application---commonly capturing local map information in \gls{a:av} applications and robotic systems.}
  \item \added[id=gf]{We provide a set of \gls{k:offline} and \gls{k:online} demonstrations of \gls{a:strem} through (i) the \gls{k:dataset} dataset and (ii) \gls{a:carla} simulator with \gls{a:ros} integration, respectively.}
\end{itemize}
The aforementioned contributions are in addition to the contributions of our preliminary work \cite{anderson2023pattern}, which comprised: 
\begin{itemize}
  \item The \gls{s:rexs4u} querying language (and associated semantics) for pattern matching spatio-temporal sequences of events in perception streams.
  \item The \gls{a:strem} tool for \gls{k:offline} and \gls{k:online} pattern matching of perception streams with the novel querying language.
  \item An \gls{k:offline} demonstration of the \gls{a:strem} tool using the \gls{k:dataset} \gls{a:av} dataset.
  \item An \gls{k:online} demonstration of \gls{a:strem} through integration with the \gls{a:ros} and \gls{a:carla} simulator stack.
\end{itemize}


%% file: src/sections/preliminaries.tex
\section{Preliminaries}
\label{sec:prelims}

We let \gls{s:integers} be the set of all integers, \gls{s:naturals}, $\gls{s:naturals}_{0}$ be the set of natural numbers with and without 0\added{, respectively}, and \gls{s:reals} be the set of real numbers.
Furthermore, \gls{s:booleans} represents the set of booleans $\{\gls{s:true}, \gls{s:false}\}$ where \gls{s:true} and \gls{s:false} are the boolean constants true and false, respectively.
Furthermore, given a set $A$, $\gls{s:powerset}(A)$ denotes the powerset of $A$, and \cardinality{A} represents the cardinality of $A$.

\input{src/sections/preliminaries/stream}


%% file: src/sections/preliminaries/stream.tex
\subsection{Perception Stream}
\label{sec:prelims:stream}

\begin{figure*}
    \centering
    \includegraphics[width=1.0\linewidth]{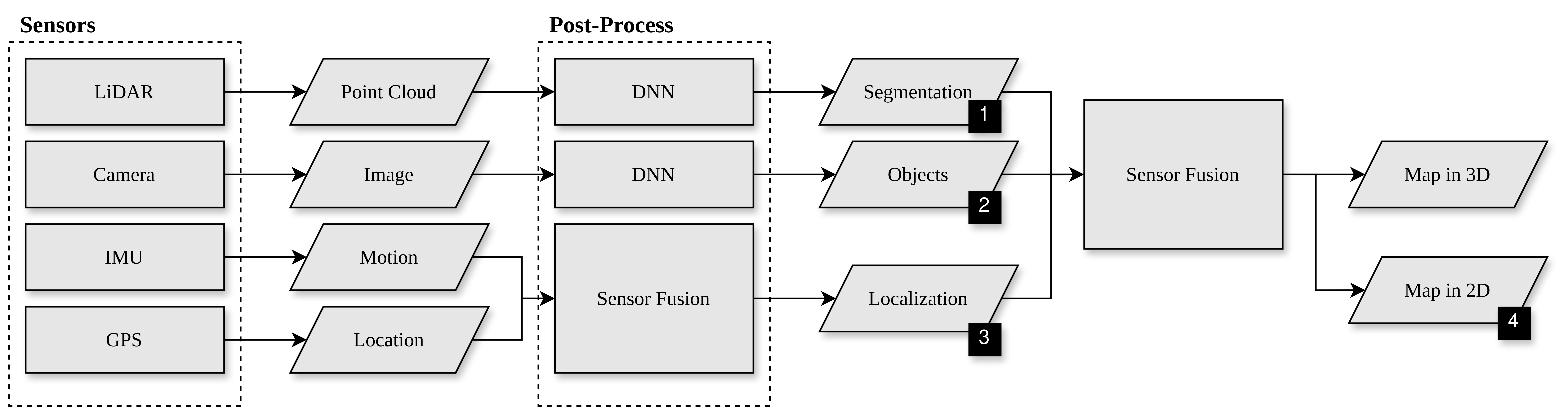}
    \caption{\added[id=gf]{An example of an \glsfmttext{a:av} perception stack pipeline. In our running examples and case studies, the perception data may be received from: (1) segmentation data from \glsfmttext{a:lidar}, (2) object annotations from images, (3) geo-spatial data, or (4) localization maps.}}
    \label{fig:prelims:stream:0}
\end{figure*}

We consider a perception stream $\gls{s:datastream} = \gls{s:frame}_{0}, \gls{s:frame}_{1}, \gls{s:frame}_{2}, \gls{s:frame}_{3}, \ldots$ to be a discrete sequence of frames $\gls{s:frame}_{i}$ where $i \in \gls{s:naturals}_{0}$ represents the $i$\textsuperscript{th} frame of the stream.
We use \gls{s:subseq} to denote the subsequence relation, i.e., $\gls{s:datastream}'\: \gls{s:subseq}\: \gls{s:datastream}$ where $\gls{s:datastream}' = (\gls{s:frame}_{i}, \ldots, \gls{s:frame}_{j})$ is a subsequence of \gls{s:datastream} \added{such that $i \leq j$, and $j < \cardinality{\gls{s:datastream}}$}.
For readability, we henceforth refer to a subsequence $\gls{s:datastream}'$ of \gls{s:datastream} as a range of frames using the shorthand notation $\gls{s:frame}_{i, j}$ where $i \leq j$.
\added{In the case that $i = j$ for some frame $\gls{s:frame}_{i, j}$, we use $\gls{s:frame}_{i}$ to denote a single frame from the perception stream \gls{s:datastream} such that $i < \cardinality{\gls{s:datastream}}$.}
\added{Furthermore, the data associated with a perception stream is highly system-dependent and takes various forms in order to effectively perceive the environment.}
\added{For instance, consider \cref{fig:prelims:stream:0} of an \gls{a:av} perception stack that utilizes multiples sensors, modalities, and sensor fusion technologies to perceive its environment.}
\added{From this perception stack, perception data may come from different points along the pipeline depending on what needs to be searched over, accordingly.}
\added[id=gf]{For example, when the perception stream source is the data stream $\boxed{2}$ in \cref{fig:prelims:stream:0} (image object annotations channel), a perception stream \gls{s:datastream} with three frames $\gls{s:frame}_{0,2}$  is visualized in \cref{fig:prelims:stream:1}.}

Each frame $\gls{s:frame}_{i}$ in the perception stream is \added{assumed to be} a \gls{k:key-frame} \replaced{that contains a series of channels \gls{s:channels} comprised of data sampled from various sensors and sensor fusions}{that contains data generated from one or more sensor channels \gls{s:channels}}.
A \gls{k:key-frame} is a frame where the timestamp difference between \added{the sensor data sampled from each channel in \gls{s:channels}} is within some threshold $\gls{s:tsthresh} \in \gls{s:reals}$.
For example, \replaced{consider a perception system}{a frame} of an \gls{a:av} affixed with a front, front-left, front-right, and rear channels \added{associated with monocular camera sensors}\added{; a frame is only labeled as} a \gls{k:key-frame} if and only if the timestamp difference between all \added{channel} samples occur within a period of 0.001s.
Furthermore, each frame contains a finite set of object annotations (henceforth, ``objects'') $O$ from the entire stream \gls{s:datastream} of objects \gls{s:objects} consisting of query-able information.
That is, each object $o \in \gls{s:objects}$ may be annotated with a label (e.g., \textit{car}, \textit{pedestrian}, \textit{sign}, \textit{bus}, \textit{animal}, etc.), a unique identifier (``ID''), 2D/3D bounding box, confidence score, segmentation map, \gls{a:lidar} points, etc.

\begin{figure*}[thb]
  \centering
  \begin{tikzpicture}
    \node[draw,fill=white,drop shadow=black,rectangle,minimum width=3cm,minimum height=3cm,anchor=north west] (f1) {};
    \node[above=0cm of f1.south east,xshift=-0.25cm,anchor=south] (label1) {$\replaced{\protect\ensuremath{\gls{s:environment}}}{\glsfmttext{s:imagespace}}_{c}$};
    \node[yshift=-0.3cm] at (f1.south) {$\glsfmttext{s:frame}_{0}$};

    \node[draw,red,rectangle,minimum width=1.7cm,minimum height=0.8cm,anchor=north west,xshift=0.2cm,yshift=-0.2cm] (bbox1) at (f1.north west) {};
    \node[draw,red,fill=red,text=white,rectangle,minimum width=1.0cm,minimum height=0.1cm,anchor=north west] (label1) at (bbox1.south west) {\scriptsize bus};
    
    \node[draw,red,rectangle,minimum width=1.5cm,minimum height=1.3cm,anchor=north west,xshift=1cm,yshift=-0.4cm] (bbox2) at (f1.north west) {};
    \node[draw,red,fill=red,text=white,rectangle,minimum width=1.0cm,minimum height=0.1cm,anchor=north west] (label2) at (bbox2.south west) {\scriptsize bus};

    \draw [black,line width=0.5pt,-stealth,shorten >=-0.5mm,shorten <=-0.25pt](f1.north west) -- (f1.north east);
    \draw [black,line width=0.5pt,-stealth,shorten >=-0.5mm,shorten <=-0.25pt](f1.north west) -- (f1.south west);

    \node[draw,fill=white,drop shadow=black,rectangle,minimum width=3cm,minimum height=3cm,anchor=north west,right=of f1] (f2) {};
    \node[above=0cm of f2.south east,xshift=-0.25cm,anchor=south] (label1) {$\replaced{\protect\ensuremath{\gls{s:environment}}}{\glsfmttext{s:imagespace}}_{c}$};
    \node[yshift=-0.3cm] at (f2.south) {$\glsfmttext{s:frame}_{1}$};
    
    \node[draw,red,rectangle,minimum width=1.7cm,minimum height=0.8cm,anchor=north west,xshift=0.4cm,yshift=-0.2cm] (bbox1) at (f2.north west) {};
    \node[draw,red,fill=red,text=white,rectangle,minimum width=1.0cm,minimum height=0.1cm,anchor=north west] (label1) at (bbox1.south west) {\scriptsize bus};
    
    \node[draw,blue,rectangle,minimum width=1.5cm,minimum height=1.3cm,anchor=north west,xshift=1.2cm,yshift=-0.5cm] (bbox2) at (f2.north west) {};
    \node[draw,blue,fill=blue,text=white,rectangle,minimum width=1.0cm,minimum height=0.1cm,anchor=north west] (label2) at (bbox2.south west) {\scriptsize ped};

    \node[draw,green,rectangle,minimum width=2cm,minimum height=0.8cm,anchor=north west,xshift=0.5cm,yshift=-1.6cm] (bbox2) at (f2.north west) {};
    \node[draw,green,fill=green,text=white,rectangle,minimum width=1.0cm,minimum height=0.1cm,anchor=north west] (label2) at (bbox2.south west) {\scriptsize car};
    
    \draw [black,line width=0.5pt,-stealth,shorten >=-0.5mm,shorten <=-0.25pt](f2.north west) -- (f2.north east);
    \draw [black,line width=0.5pt,-stealth,shorten >=-0.5mm,shorten <=-0.25pt](f2.north west) -- (f2.south west);

    \node[draw,fill=white,drop shadow=black,rectangle,minimum width=3cm,minimum height=3cm,anchor=north west,right=of f2] (f3) {};
    \node[above=0cm of f3.south east,xshift=-0.25cm,anchor=south] (label1) {$\replaced{\protect\ensuremath{\gls{s:environment}}}{\glsfmttext{s:imagespace}}_{c}$};
    \node[yshift=-0.3cm] at (f3.south) {$\glsfmttext{s:frame}_{2}$};

    \node[draw,red,rectangle,minimum width=1.7cm,minimum height=0.8cm,anchor=north west,xshift=0.6cm,yshift=-0.2cm] (bbox1) at (f3.north west) {};
    \node[draw,red,fill=red,text=white,rectangle,minimum width=1.0cm,minimum height=0.1cm,anchor=north west] (label1) at (bbox1.south west) {\scriptsize bus};
    
    \node[draw,blue,rectangle,minimum width=1.5cm,minimum height=1.3cm,anchor=north west,xshift=1.2cm,yshift=-0.7cm] (bbox2) at (f3.north west) {};
    \node[draw,blue,fill=blue,text=white,rectangle,minimum width=1.0cm,minimum height=0.1cm,anchor=north west] (label2) at (bbox2.south west) {\scriptsize ped};
    
    \draw [black,line width=0.5pt,-stealth,shorten >=-0.5mm,shorten <=-0.25pt](f3.north west) -- (f3.north east);
    \draw [black,line width=0.5pt,-stealth,shorten >=-0.5mm,shorten <=-0.25pt](f3.north west) -- (f3.south west);
  \end{tikzpicture}
  \vspace{-0.3em}\caption{An example perception stream \glsfmttext{s:datastream}, \added{sourced from (2) in \cref{fig:prelims:stream:0}}, containing the frames $\gls{s:frame}_{0}, \gls{s:frame}_{1}, \gls{s:frame}_{2}$ of a camera sensor channel \replaced{$\gls{s:channel} \in \gls{s:channels}$}{$c$} with \replaced{image}{pixel} space $\replaced{\protect\ensuremath{\gls{s:environment}}}{\gls{s:imagespace}}_{\replaced{\gls{s:channel}}{c}}\added{\protect\ensuremath{\subseteq \gls{s:naturals}}}$. For each object in a given frame, a classification and bounding box is minimally assumed to be \replaced{attributed}{annotated}. In addition, each frame may be augmented with other sensor data relevant to the system to provide further context such as \replaced{\gls{a:gps}, \gls{a:imu}}{GPS, IMU}, etc.}
  \label{fig:prelims:stream:1}
\end{figure*}

\begin{figure*}[thb]
  \centering
  \begin{tikzpicture}
    \node[draw,inner sep=0,fill=white,drop shadow=black,rectangle,minimum width=3cm,minimum height=3cm,anchor=north west] (f1) {\includegraphics[width=3.5cm]{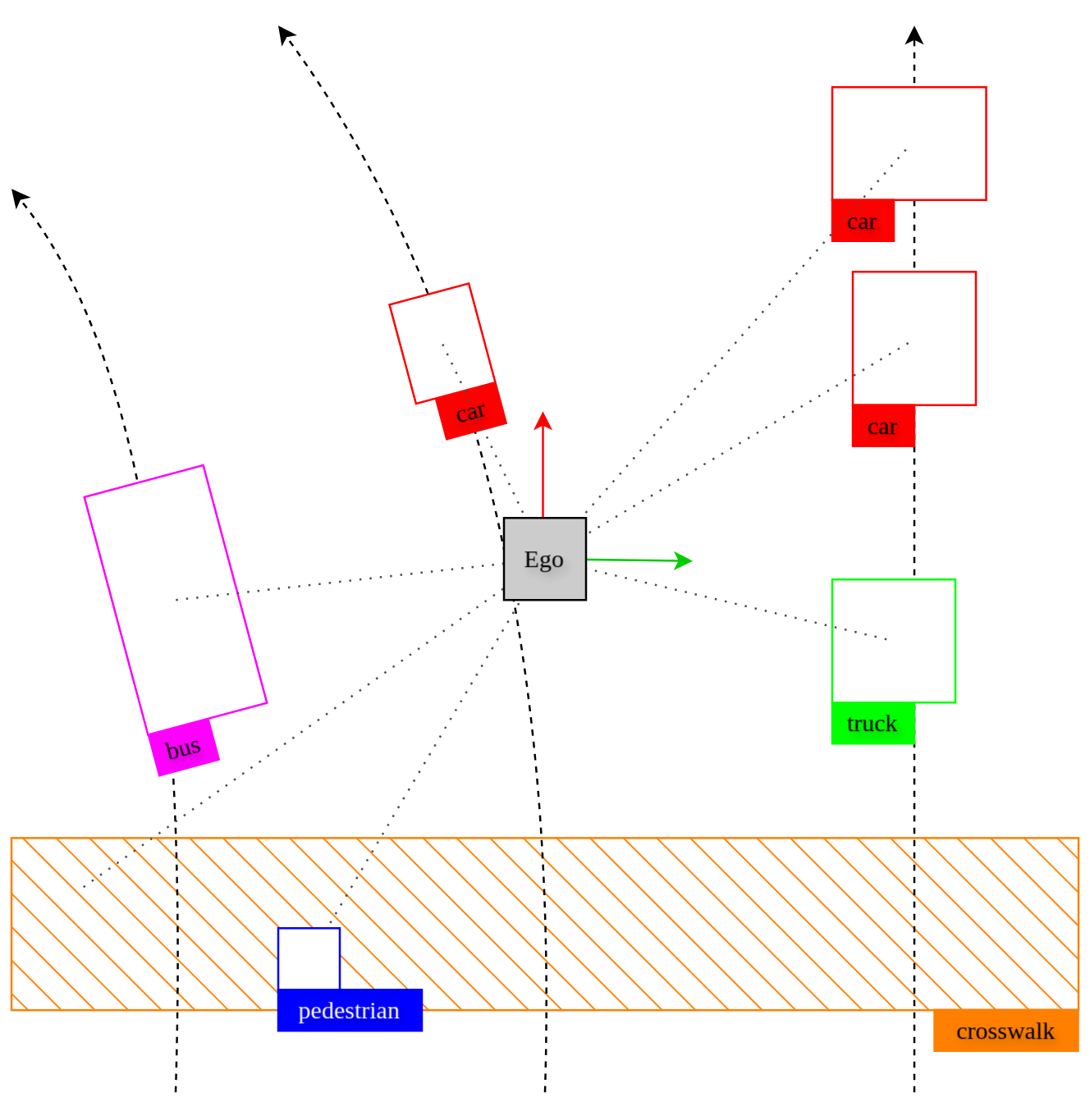}};
    \node[above=0cm of f1.south east,xshift=-0.25cm,anchor=south] (label1) {$\glsfmttext{s:environment}_{\glsfmttext{s:channel}}$};
    \node[yshift=-0.3cm] at (f1.south) {$\glsfmttext{s:frame}_{0}$};


    \node[draw,inner sep=0,fill=white,drop shadow=black,rectangle,minimum width=3cm,minimum height=3cm,anchor=north west,right=of f1] (f2) {\includegraphics[width=3.5cm]{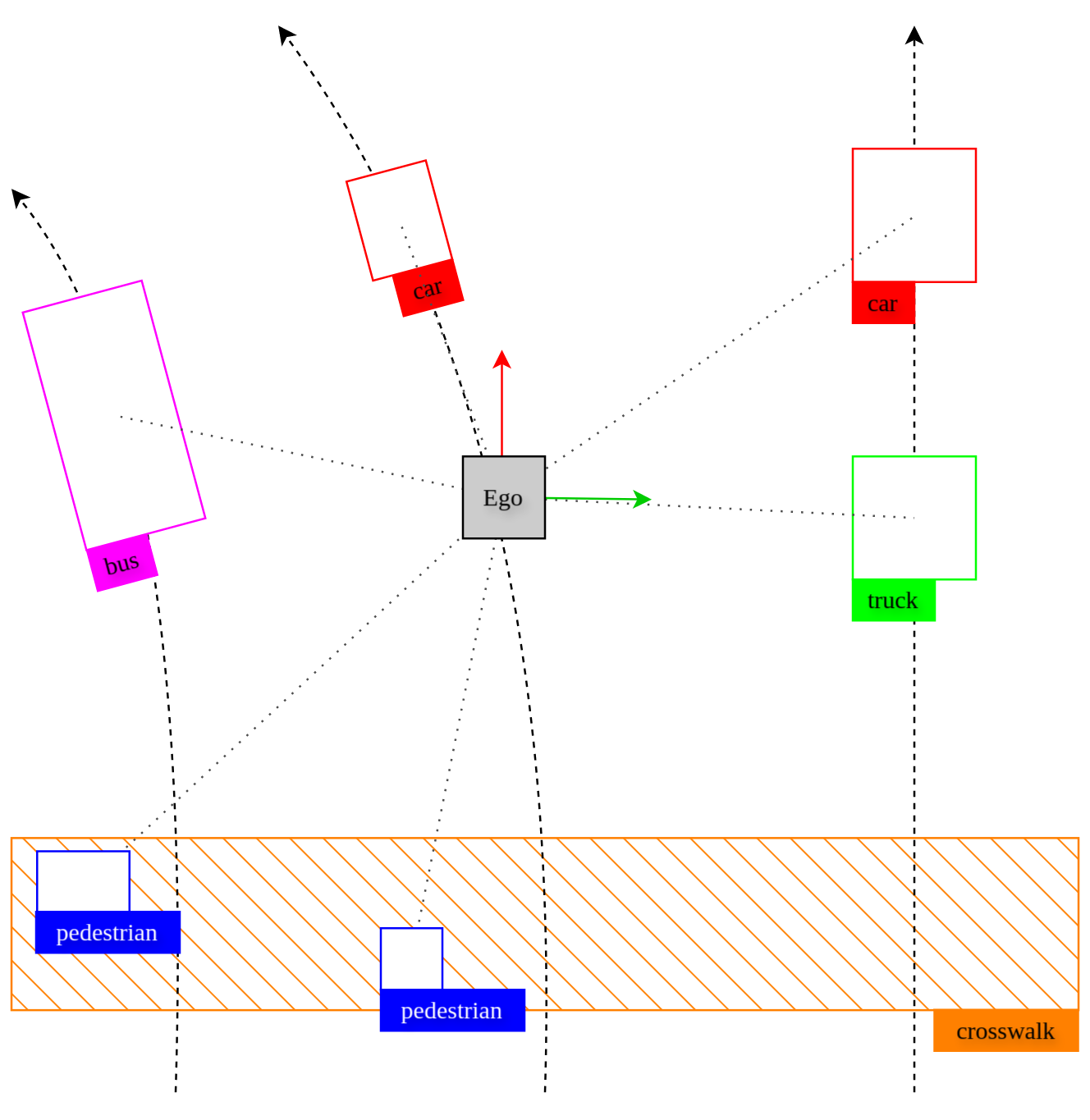}};
    \node[above=0cm of f2.south east,xshift=-0.25cm,anchor=south] (label1) {$\glsfmttext{s:environment}_{\glsfmttext{s:channel}}$};
    \node[yshift=-0.3cm] at (f2.south) {$\glsfmttext{s:frame}_{1}$};
    

    \node[draw,inner sep=0,fill=white,drop shadow=black,rectangle,minimum width=3cm,minimum height=3cm,anchor=north west,right=of f2] (f3) {\includegraphics[width=3.5cm]{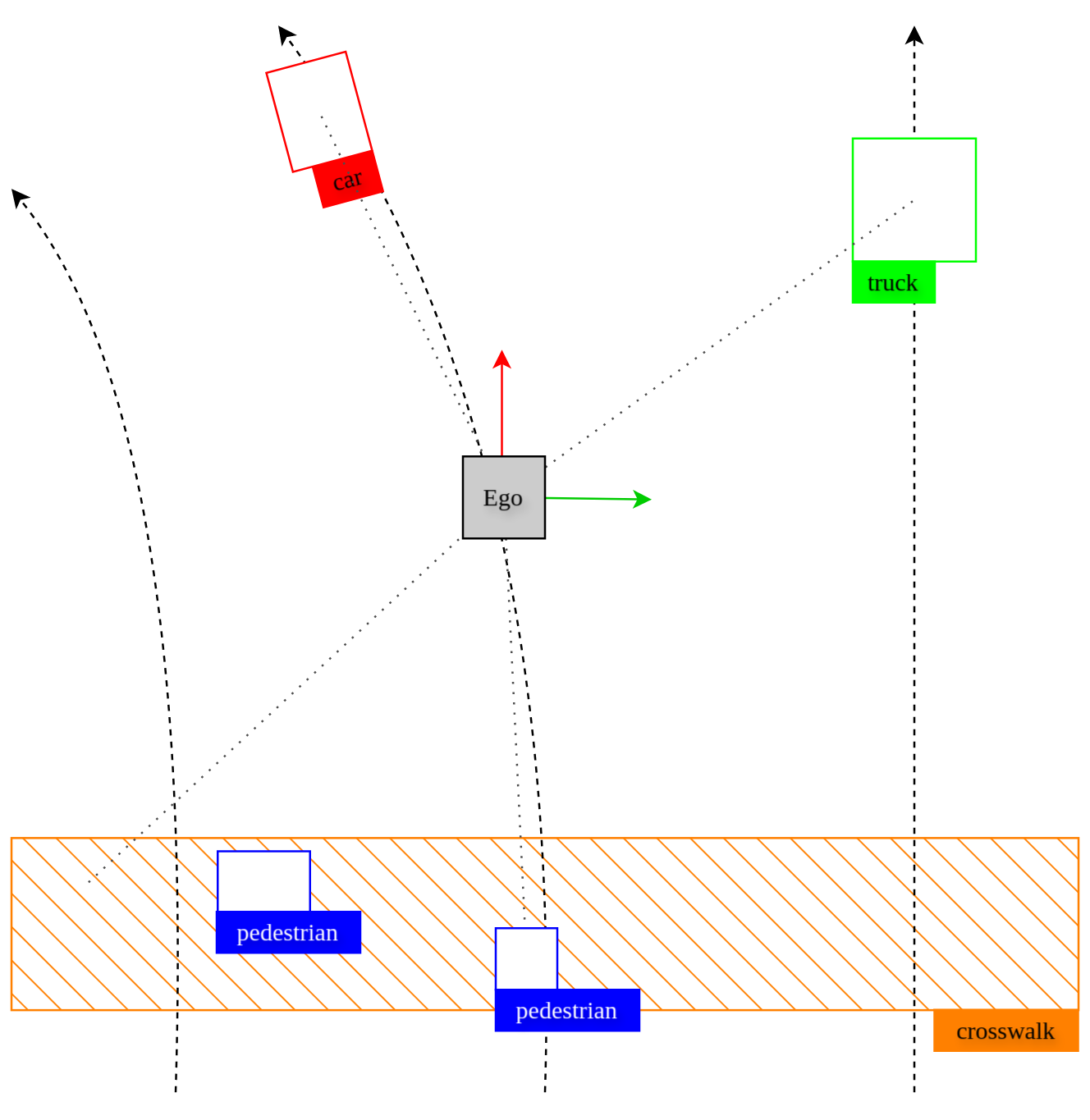}};
    \node[above=0cm of f3.south east,xshift=-0.25cm,anchor=south] (label1) {$\glsfmttext{s:environment}_{\glsfmttext{s:channel}}$};
    \node[yshift=-0.3cm] at (f3.south) {$\glsfmttext{s:frame}_{2}$};
    
  \end{tikzpicture}
  \caption{\added{An example perception stream \gls{s:datastream}, sourced from (4) in \cref{fig:prelims:stream:0}, containing the frames $\gls{s:frame}_{0}$, $\gls{s:frame}_{1}$, $\gls{s:frame}_{2}$ of a \gls{k:sensor-fusion} channel $\gls{s:channel} \in \gls{s:channels}$ with \gls{k:top-down} space $\gls{s:environment}_{c} \subseteq \gls{s:reals}^{3}$. For each object in a given frame, a classification, rotation, and bounding box is minimally assumed to be attributed. In addition, the \gls{k:top-down} view may enable additional sensor information such as \textit{lanes}, \textit{crosswalks}, or \textit{bike lanes}.}}
  \label{fig:prelims:stream:2}
\end{figure*}

In the following, we assume that given an object $o \in \gls{s:objects}$, we have defined functions that return the required annotation and/or auxiliary data.
For example, for retrieving qualitative attributes, we define the function $\gls{s:attrs} : \gls{s:objects} \times \gls{a:keys} \rightarrow \gls{s:attributes}$
where $\gls{a:keys}$ is a set of keys and $\gls{s:attributes}$ is a set of attributes.
We will not formally define the sets $\gls{a:keys}$ and $\gls{s:attributes}$ since they are \replaced{system}{dataset} dependent, but as an example, we could have \replaced{\textit{class}}{{\tt class}} and \replaced{\textit{color}}{{\tt color}} as keys, and \replaced{\textit{bus}}{{\tt bus}} and \replaced{\textit{green}}{{\tt green}} as the corresponding attributes\added{, respectively}.
\added[id=gf]{Moreover, for a given set of objects $O \in \gls{s:objects}$, we define the function 
\[ \gls{s:objfn}(\gls{s:annot}, O) = \lbrace o \in O \mid \forall{a \in \gls{s:annot}.\: \exists{k \in \gls{a:keys}.\: \gls{s:attrs}(o, k) = a}} \rbrace\]
as an abbreviation to retrieve the set of objects that match a specified set of attributes \gls{s:annot}---see \cref{def:spre:syntax:1,def:spre:syntax:2} for further discussion of \gls{s:annot}.}
When formulating a query, we are searching to find in a frame an object that satisfies certain attributes, e.g., a \textit{green bus}.
\added{In addition, for retrieving spatial-based attributes (e.g., bounding boxes or regions) of objects, the function $\gls{s:bbox}_{c} : \gls{s:objects} \rightarrow \gls{s:powerset}(\gls{s:environment}_{c})$ is used where $\gls{s:environment}_{c}$ represents the working environment contextualized to a particular channel $c \in \gls{s:channels}$.}
\added{This working environment may be, for example, the image space of a camera (see \cref{fig:prelims:stream:1}) or a top-down perspective of the surrounding environment from sensor fusion or simulation-based sampling methods (see \cref{fig:prelims:stream:2}).}
\cref{tab:prelims:stream:1} presents a detailed layout of \cref{fig:prelims:stream:1} with annotated information for each object \added{from a single channel of the perception system}.

\begin{table*}[thb]
  \centering
  \caption{An example of annotated data from a perception stream. Each object has some attributes along with some bounding box (``BB'') data. Depending on the application, the object IDs may be unique across the stream, unique only in each frame, or unique to each frame for each class of objects. In addition, each object may be annotated with additional data such as \glsxtrshort{a:lidar} points, color (if applicable), etc.}
  \begin{tabular*}{1.0\linewidth}{@{\extracolsep{\fill}}ccc}
    \toprule
    $\gls{s:frame}_{0}$ & $\gls{s:frame}_{1}$ & $\gls{s:frame}_{2}$\\
    \midrule
    (bus, red, ID: 1, BB) & (bus, red, ID: 1, BB) & (bus, red, ID: 1, BB)\\
    (bus, yellow, ID: 2, BB) & (pedestrian, child, ID: 2, BB) & (pedestrian, child, ID: 2, BB)\\
    & (car, sedan, ID: 3, BB) &\\
    \bottomrule
  \end{tabular*}
  \label{tab:prelims:stream:1}
\end{table*}

%% file: src/sections/spre.tex
\section{\glsfmtlongpl{a:spre}}
\label{sec:spre}

\gls{a:spre} (pronounced /spri:/) is a querying language designed to capture scenarios of perception streams.
The \gls{s:rexs4u} language leverages the power of pattern matching from \glspl{a:re} \cite{aho1980pattern} with the topological reasoning of \gls{s:s4u} logic \cite{kontchakov2007spatial}.
The practice of merging a formal logic with an \gls{a:re}-based language to produce an extended, more expressive, version has previously been studied in \cite{wolper1983temporal,beer2001temporal}.
However, these efforts primarily focus on extending temporal-based logics such as \gls{a:ltl} \cite{pnueli1977temporal} with \glspl{a:re}, whereas current extensions to spatial-based logics with \glspl{a:re} is unheard of to the best of our knowledge---particularly, the \gls{s:s4u} branch of logic.
Thus, the \gls{a:spre} language aims to provide a formal approach in extending pattern-based constructs with spatial-based formulas.

Within this section, we first introduce the formal syntax of the \gls{a:spre} language followed by its semantics interpreted over perception streams.

\input{src/sections/spre/syntax}
\input{src/sections/spre/semantics}


%% file: src/sections/spre/syntax.tex
\subsection{Syntax}
\label{sec:spre:syntax}

The \gls{a:spre} syntax consists of \replaced{four}{three} interdependent grammars that make up the querying language\added{: (1) \gls{s:s4+} (``\gls{k:spatial-term}''), (2) \gls{s:s4m+} (``\gls{k:metric-expression}''), (3) \gls{s:s4u+} (``\gls{k:spatial-formula}''), and (4) \gls{a:spre} (``\gls{k:query}'')}.
\added{The first three grammars are inspired by the \gls{s:s4} and \gls{s:s4u} modal logics for reasoning over topological spaces; and the last grammar is inspired by traditional \glspl{a:re} commonly used in \gls{k:pattern-matching} problems.}

\added{We introduce each component of the \gls{a:spre} language separately as the interpretation (i.e., semantics) of each is unique to its respective grammar.}
\added{Moreover, although the \gls{s:s4+} and \gls{s:s4u+} grammars have an identical syntax to their counterparts, \gls{s:s4} and \gls{s:s4u}, respectively, we re-introduce them here as new semantics for these logics will be defined.}

\subsubsection{\glsentrytitlecase{k:spatial-term}{name}}
\label{sec:spre:syntax:terms}

\added{A \gls{k:spatial-term} enables set operations over set-based attributes of objects (e.g., bounding boxes).}

\begin{definition}[$\bm{\glsentryname{s:s4+}}$ Syntax]
    \label{def:spre:syntax:1}
    The structure of an \gls{s:s4+} formula \gls{s:s4+form} is inductively defined by the following grammar:
    \begin{equation*}
        \gls{s:s4+form} ::= \gls{s:annot} \mid
        \added{\protect\ensuremath{v \mid}}
        \bar{\gls{s:s4+form}} \mid
        \gls{s:s4+form}_{1}\: \gls{s:inter}\: \gls{s:s4+form}_{2} \mid
        \gls{s:s4+form}_{1}\: \gls{s:union}\: \gls{s:s4+form}_{2} \mid
        \gls{s:interior}\: \gls{s:s4+form} \mid
        \gls{s:closure}\: \gls{s:s4+form}
    \end{equation*}
    where \gls{s:annot} is an atomic proposition ranging over sets of attributes $\gls{s:powerset}(\gls{s:attributes})$; \added[id=gf]{$v \in \gls{s:vars}$ is a variable; }
    $\bar{\gls{s:s4+form}}$ is the unary operator for complement; 
    \gls{s:inter} and \gls{s:union} are the binary operators for intersection and union; and \gls{s:interior} and \gls{s:closure} are the interior and closure operators.
\end{definition}

    \added{Informally, we introduce \gls{s:vars} as a set of variables within the scope of an \gls{s:s4+} formula.}
    \added{The introduction of variables at this level of the grammar supports quantifying over a set of objects within a frame in order to perform multiple set operations on the same object within a single \gls{s:s4+} or \gls{s:s4m+} formula.}
    \added{For instance, the variable $v \in \gls{s:vars}$ may resolve to a \textit{car} in the scene that is \textit{blue}, and we may want to retrieve both the interior and the complement of the same object within a single \gls{s:s4+} formula.}


\subsubsection{\glsentrytitlecase{k:metric-expression}{name}}
\label{sec:spre:syntax:metric}

\added[id=gf]{A \gls{k:metric-expression} enables arbitrary arithmetic-based operations over numerical interpretations of the underlying spatial terms.}

\begin{definition}[\added{$\bm{\glsentryname{s:s4m+}}$ Syntax}]
  \label{def:spre:syntax:2}
  \added{Given the \glspl{k:spatial-term} \gls{s:s4+form}, $\gls{s:s4+form}_{1}$, and $\gls{s:s4+form}_{2}$ the structure of an \gls{s:s4m+} formula \gls{s:s4m+form} is inductively defined by the following grammar:}
  \begin{align*}
      \added{\protect\ensuremath{\gls{s:s4m+form} ::=}} &
      \added{\protect\ensuremath{c \mid}}
      \added{\protect\ensuremath{f(\gls{s:s4+form}) \mid}}
      \added{\protect\ensuremath{g(\gls{s:s4+form}_{1}, \gls{s:s4+form}_{2}) \mid}}
      \added{\protect\ensuremath{- \gls{s:s4m+form} \mid}}
      \added{\protect\ensuremath{\gls{s:s4m+form}_{1} + \gls{s:s4m+form}_{2} \mid}}\\
      & \added{\protect\ensuremath{\gls{s:s4m+form}_{1} * \gls{s:s4m+form}_{2} \mid}}
      \added{\protect\ensuremath{\gls{s:s4m+form}^{c}}}
  \end{align*}
 \added{where $c \in \gls{s:reals}$; $f$ and $g$ are generic unary/binary functions that return some value in \gls{s:reals}, and $+$, $*$, and $\gls{s:s4m+form}^{c}$ are the standard arithmetic operations addition, multiplication, and exponentiation, respectively}.
\end{definition}

\added{The functions $f$ and $g$ operate on \glspl{k:spatial-term} and may compute various features from the objects in a particular working environment such as the area, volume, distance, \gls{a:iou}, etc.}

\subsubsection{\glsentrytitlecase{k:spatial-formula}{name}}
\label{sec:spre:syntax:formulas}

\added{A \gls{k:spatial-formula} extends \glspl{k:spatial-term} and \glspl{k:metric-expression} by enabling boolean-based valuations (e.g., relational operations, non-emptiness checks, etc).}

\begin{definition}[$\bm{\glsentryname{s:s4u+}}$ Syntax]
    \label{def:spre:syntax:3}
    Given the \glspl{k:spatial-term} \gls{s:s4+form}, $\gls{s:s4+form}_{1}$, and $\gls{s:s4+form}_{2}$, the metric expressions $\gls{s:s4m+form}_{1}$ and $\gls{s:s4m+form}_{2}$\added{, and a variable $v \in V$,} the structure of an \gls{s:s4u+} formula \gls{s:s4u+form} is inductively defined by the following grammar:
    \begin{align*}
        \gls{s:s4u+form} ::=& \gls{s:annot} \mid
        \added[id=gf]{\protect\ensuremath{\exists{v}(\gls{s:annot}).\gls{s:s4u+form} \mid}}
        \gls{s:isnonempty}\: \gls{s:s4+form} \mid
        \gls{s:s4+form}_{1}\: \gls{s:issubset}\: \gls{s:s4+form}_{2} \mid
        \gls{s:not}\: \gls{s:s4u+form} \mid
        \gls{s:s4u+form}\: \gls{s:and}\: \gls{s:s4u+form} \mid\\
        & \gls{s:s4u+form}\: \gls{s:or}\: \gls{s:s4u+form} \mid
        \gls{s:s4m+form}_{1} \leq \gls{s:s4m+form}_{2}
    \end{align*}
    where \gls{s:annot} is an atomic proposition ranging over sets of attributes $\gls{s:powerset}(\gls{s:attributes})$; \added{$\exists{v}(\gls{s:annot})$ is the existential quantification operator;} \gls{s:isnonempty} and \gls{s:issubset} are the boolean operators for set emptiness and set inclusion; \deleted{and} \gls{s:not}, \gls{s:and}, and \gls{s:or} are the standard propositional logic operators\added{; and $\leq$ is the inequality operator over the reals}.
\end{definition}

    \added[id=gf]{In natural language, the expression $\exists{v}(\gls{s:annot}).\gls{s:s4u+form}$ should be read as ``there exists an object with attributes $\gls{s:annot}$ (stored in variable $v$) that satisfies the formula $\gls{s:s4u+form}$".}
    \added{We note that with the introduction of the quantification operator, we consider a spatial formula \gls{s:s4u+form} to be a \gls{a:wff} if and only if there are no free variables.}
    \added[id=gf]{That is, we require that all the variables that appear in a formula to be within the scope of an existential operator.}
    \added[id=gf]{For instance, the formula $\exists{v(\gls{s:annot}_1).(v\: \gls{s:and}\: \gls{s:annot}_2)}$ is a \gls{a:wff} as the variable $v$ is used within the scope of  $\exists{v(\gls{s:annot})}$.}
    \added[id=gf]{On the other hand, the formula $\exists{u(\gls{s:annot}).(v\: \gls{s:and}\: u)}$ is not \gls{a:wff} since the variable $u$ does not appear under the scope of an existential quantifier for $u$}.

\subsubsection{\glsentrytitlecase{k:query}{name}}
\label{sec:spre:syntax:query}

\added{A \gls{k:query} (interchangeably, \gls{a:spre}) extends \glspl{k:spatial-formula} by enabling \gls{a:re}-based constructs to express patterns over spatial information.}

\begin{definition}[\glsxtrshort{a:spre} Syntax]
    \label{def:spre:syntax:4}
    Given the \gls{k:spatial-formula} \gls{s:s4u+form}, the structure of a \glsxtrshort{a:spre} query is inductively defined by the following grammar:
    \begin{equation*}
        \gls{s:query} ::= \gls{s:s4u+form} \mid
        \gls{s:query}_{1}\ \gls{s:alt}\ \gls{s:query}_{2} \mid
        \gls{s:query}_{1}\: \gls{s:concat}\: \gls{s:query}_{2} \mid
        \gls{s:query}{\gls{s:kleene}}
    \end{equation*}
    where the operators \gls{s:alt}, \gls{s:concat}, and $\phantom{}^{\gls{s:kleene}}$ are the standard \gls{a:re} operations alternation, concatenation, and Kleene-star, respectively.
\end{definition}

\begin{remark}[Alphabet]
    We consider the alphabet $\gls{s:alphabet} = \lbrace \gls{s:symbol}_{1}, \gls{s:symbol}_{2}, ..., \gls{s:symbol}_{n} \rbrace = \gls{s:powerset}({\gls{s:objects}})$ where each symbol represents a possible combination of the objects from the perception stream \gls{s:datastream}.
    The main intuition is that for every frame of the stream of perception data, we would like to match a symbol with only the relevant objects for the given \gls{a:spre} query---see \cref{ex:spre:semantics:1} below.
\end{remark}

\begin{example}
    \label{ex:spre:semantics:1}
    Consider the data presented in Table \ref{tab:prelims:stream:1}. 
    The alphabet \gls{s:alphabet} will contain 64 symbols in total. 
    Some examples of symbols from \gls{s:alphabet} could be:
    \begin{align*}
        \gls{s:symbol}_{1} &= \textnormal{\{(bus, red, ID: 1, BB)\}}\\
        \gls{s:symbol}_{2} &= \begin{aligned}[t]
            \textnormal{\{} & \textnormal{(pedestrian, child, ID: 2, BB),}\\
            & \textnormal{(bus, yellow, ID: 2, BB)} \}
        \end{aligned} \\
        \gls{s:symbol} _{3} &= \begin{aligned}[t]
            \textnormal{\{} & \textnormal{(bus, red, ID: 1, BB),}\\
            & \textnormal{(car, sedan, ID: 3, BB)\}}
        \end{aligned}
    \end{align*}
    When we query for \textit{bus}, we would like our pattern matching algorithm to return $\gls{s:symbol} _{1}$ in the matching strings, but not $\gls{s:symbol} _{2}$ or $\gls{s:symbol} _{3}$.
\end{example}


%% file: src/sections/spre/semantics.tex
\subsection{Semantics}
\label{sec:spre:semantics}

\replaced{\Gls{k:pattern-matching}}{Pattern matching} on perception streams differs from the standard string pattern matching \added{problem over strings}.
When querying perception data, the goal is to identify annotations represented as abstract objects affixed with some attributes.
For example, a query could be ``\textit{Find all sequences where a car appears in at least three consecutive frames}''.
In such a query, we do not ask for a specific car with a unique identity (which is not known in advance), but rather for any car.
In addition, there may be multiple cars in a frame which implies that all of them should be candidates for a pattern match.
In other words, even though our patterns in \glspl{a:spre} are over object attributes, our queries should return strings where each symbol is a set of corresponding specific objects.

We define the semantics of \gls{s:s4+} \replaced{formulas}{expressions} through \replaced{the following}{a} valuation function\added{:} $\gls{s:valuation}{\;}: \gls{s:powerset}(\gls{s:objects}) \to \gls{s:powerset}({\gls{s:powerset}({\replaced{\protect\ensuremath{\gls{s:environment}_{c}}}{\protect\ensuremath{\gls{s:reasonspace}}}))}}$\deleted{$W$ is the spatial reasoning space that resolves to either the camera sensor's image space $\gls{s:imagespace}_{\gls{s:channel}}$ (i.e., pixels) or the working environment \gls{s:environment}, } where $\gls{s:powerset}(\replaced{\protect\ensuremath{\gls{s:environment}_{c}}}{\gls{s:reasonspace}})$ is all possible bounding boxes from the reasoning space, and $\gls{s:powerset}(\gls{s:powerset}(\replaced{\protect\ensuremath{\gls{s:environment}_{c}}}{\gls{s:reasonspace}}))$ is the set of all possible sets of all bounding boxes.
The spatial terms of \added{an} \gls{s:s4+} \added{formula} specify set-theoretic operations over bounding boxes of objects from the perception stream \gls{s:datastream}.
\added{The semantics of a \gls{k:metric-expression} from an \gls{s:s4m+} formula provide geometric-based arithmetic operations interpreted over sets represented as bounding boxes.}
We define the semantics of \gls{s:s4u+} using a boolean satisfaction relation since our goal is to determine whether certain relations are true or not over bounding boxes and other object annotations.

\begin{definition}[$\bm{\glsentryname{s:s4+}}$ Semantics]
    \label{def:spre:semantics:1}
    Given a set of objects $O\: \gls{s:subseteq}\: \gls{s:objects}$ and \added[id=gf]{a lookup table $\gls{s:ltable} : V \rightarrow \gls{s:objects}\: \cup\: \{\gls{s:false}\}$}, the semantics of an \gls{s:s4+} formula is inductively defined as follows:
    \begin{alignat*}{1}
        \valuation{\gls{s:annot}}\replaced{\protect\ensuremath{(O, \gls{s:ltable})}}{(O)} &= \lbrace \gls{s:bbox}(o) \mid \replaced{\protect\ensuremath{o \in \gls{s:objfn}(O, \gls{s:annot})}}{o \in O \ .  \ \forall a \in \gls{s:annot} \ . \ \exists  k \in \gls{a:keys} \ . \   \gls{s:attrs}(o,k) = a} \rbrace \\
        \added{\protect\ensuremath{\valuation{v}(O, \gls{s:ltable})}} &\added{\protect\ensuremath{= \lbrace \gls{s:bbox}(o) \mid o \in \gls{s:ltable}(v) \rbrace}}\\
        \valuation{\bar{\gls{s:s4+form}}}\replaced{\protect\ensuremath{(O, \gls{s:ltable})}}{(O)} &= \lbrace \bar{S} \mid S \in \valuation{\gls{s:s4+form}}\replaced{\protect\ensuremath{(O, \gls{s:ltable})}}{(O)} \rbrace\\
        \valuation{\gls{s:s4+form}_{1}\: \gls{s:inter}\: \gls{s:s4+form}_{2}}\replaced{\protect\ensuremath{(O, \gls{s:ltable})}}{\replaced{\protect\ensuremath{(O, \gls{s:ltable})}}{(O)}} &= \lbrace S_{1}\: \gls{s:sinter}\: S_{2} \mid S_{i} \in \valuation{\gls{s:s4+form}_{i}}\replaced{\protect\ensuremath{(O, \gls{s:ltable})}}{(O)} \rbrace
    \end{alignat*}
\end{definition}

Informally, given a set of objects \replaced{$O \in \gls{s:objects}$}{$O$} from a frame \added{$\gls{s:frame}_{i} \in \gls{s:datastream}$}, the valuation of the spatial term  \gls{s:annot}  is the set of bounding boxes of all the objects which satisfy all the attributes in \gls{s:annot}. 

\begin{definition}[$\bm{\glsentryname{s:s4m+}}$ Semantics]
    \label{def:spre:semantics:4}
    \added{Given a set of objects $O\: \gls{s:subseteq}\: \gls{s:objects}$ \added[id=ja]{and a lookup table $\gls{s:ltable} : V \rightarrow \gls{s:objects}\: \cup\: \{\gls{s:false}\}$}, the semantics of an \gls{s:s4m+} formula is inductively defined as follows:}
    \begin{alignat*}{1}
        \added{\protect\ensuremath{\valuation{c}(O, \gls{s:ltable})}}
        &\added{\protect\ensuremath{= \lbrace c \rbrace}}\\
        \added{\protect\ensuremath{\valuation{f(\gls{s:s4+form})}(O, \gls{s:ltable})}} &\added[id=ja]{\protect\ensuremath{= \lbrace f(A) \; | \;  A \in \valuation{\gls{s:s4+form}}(O, \gls{s:ltable}) \rbrace}}\\
        \added{\protect\ensuremath{\valuation{g(\gls{s:s4+form}_{1}, \gls{s:s4+form}_{2})}(O, \gls{s:ltable})}}
        &   \added[id=ja]{\protect\ensuremath{= \lbrace g(A_{1}, A_{2}) \; | \; A_{i} \in \valuation{\gls{s:s4+form}_{i}}(O, \gls{s:ltable}) \rbrace} } \\
        \added{\protect\ensuremath{\valuation{- \gls{s:s4m+form}}(O, \gls{s:ltable})}}
        &\added{\protect\ensuremath{= \lbrace -r \mid r \in \valuation{\gls{s:s4m+form}} \rbrace}(O, \gls{s:ltable})}\\
        \added{\protect\ensuremath{\valuation{\gls{s:s4m+form}_{1} + \gls{s:s4m+form}_{2}}(O, \gls{s:ltable})}}
        &\added[id=ja]{\protect\ensuremath{=}} \begin{aligned}[t]
            \added[id=ja]{\protect\ensuremath{\lbrace r_{1} + r_{2} \mid}} &\:\added[id=ja]{\protect\ensuremath{r_{1} \in \valuation{\gls{s:s4m+form}_{1}}(O, \gls{s:ltable}),}}\\
            &\: \added[id=ja]{\protect\ensuremath{r_{2} \in \valuation{\gls{s:s4m+form}_{2}}(O, \gls{s:ltable}) \rbrace}}
        \end{aligned}\\
        \added{\protect\ensuremath{\valuation{\gls{s:s4m+form}_{1} * \gls{s:s4m+form}_{2}}(O, \gls{s:ltable})}}
        &\added[id=ja]{\protect\ensuremath{=}} \begin{aligned}[t]
            \added[id=ja]{\protect\ensuremath{\lbrace r_{1} * r_{2} \mid}} &\:\added[id=ja]{\protect\ensuremath{r_{1} \in \valuation{\gls{s:s4m+form}_{1}}(O, \gls{s:ltable}),}}\\
            &\: \added[id=ja]{\protect\ensuremath{r_{2} \in \valuation{\gls{s:s4m+form}_{2}}(O, \gls{s:ltable}) \rbrace}}
        \end{aligned}\\
        \added{\protect\ensuremath{\valuation{\gls{s:s4m+form}^{c}}(O, \gls{s:ltable})}}
        &\added{\protect\ensuremath{= \lbrace r^{c} \mid r \in \valuation{\gls{s:s4m+form}}(O, \gls{s:ltable}) \rbrace}}
    \end{alignat*}
    where $r$, $r_{1}$, $r_{2}$ resolve to real numbers from \gls{s:reals}.
\end{definition}

\begin{definition}[$\bm{\glsentryname{s:s4u+}}$ Semantics]
    \label{def:spre:semantics:2}
    Given a set of objects $O\: \gls{s:subseteq}\: \gls{s:objects}$ and a lookup table $\gls{s:ltable} : V \rightarrow \gls{s:objects}\: \cup\: \{\gls{s:false}\}$, the semantics of an \gls{s:s4u+} formula is inductively defined as follows:
    \begin{alignat*}{3}
        \replaced{\protect\ensuremath{(O, \gls{s:ltable})}}{O}\ \gls{s:satis} \ & \gls{s:annot} \quad & \textnormal{iff} \quad & \replaced{\protect\ensuremath{\gls{s:objfn}(O, \gls{s:annot}) \neq \gls{s:emptyset}}}{\exists o \in O \ . \   \forall a \in \gls{s:annot} \ . \ \exists  k \in \gls{a:keys} \ . \   \gls{s:attrs}(o,k) = a} 
        \\
        \replaced{\protect\ensuremath{(O, \gls{s:ltable})}}{O}\ \gls{s:satis} \ & \gls{s:not}\: \gls{s:s4u+form} \quad & \textnormal{iff} \quad & \replaced{\protect\ensuremath{(O, \gls{s:ltable})}}{O} \not\gls{s:satis}\: \gls{s:s4u+form}
        \\
        \replaced{\protect\ensuremath{(O, \gls{s:ltable})}}{O}\ \gls{s:satis} \ & \gls{s:s4u+form}_{1}\: \gls{s:and}\: \gls{s:s4u+form}_{2} \quad & \textnormal{iff} \quad & \begin{aligned}[t]
            & \replaced{\protect\ensuremath{(O, \gls{s:ltable})}}{O} \:\gls{s:satis}\: \gls{s:s4u+form}_{1}\ \textnormal{and}\\
            & \replaced{\protect\ensuremath{(O, \gls{s:ltable})}}{O} \:\gls{s:satis}\: \gls{s:s4u+form}_{2}
        \end{aligned}
        \\
        \replaced{\protect\ensuremath{(O, \gls{s:ltable})}}{O}\ \gls{s:satis} \ & \gls{s:isnonempty}\: \gls{s:s4+form} \quad & \textnormal{iff} \quad & \exists{A \in \valuation{\gls{s:s4+form}}\replaced{\protect\ensuremath{(O, \gls{s:ltable})}}{O}.\ A \:\gls{s:neq}\: \gls{s:emptyset}}
        \\
        \replaced{\protect\ensuremath{(O, \gls{s:ltable})}}{O}\ \gls{s:satis} \ & 
        \gls{s:s4+form}_{1}\: \gls{s:issubset}\: \gls{s:s4+form}_{2} \quad & \textnormal{iff} \quad & \begin{aligned}[t]
            & \exists{A_{1} \in \valuation{\gls{s:s4+form}_{1}}\replaced{\protect\ensuremath{(O, \gls{s:ltable})}}{O}.}\\
            & \exists{A_{2} \in \valuation{\gls{s:s4+form}_{2}}\replaced{\protect\ensuremath{(O, \gls{s:ltable})}}{O}}.\\
            & A_{1} \:\gls{s:subseteq}\: A_{2}
        \end{aligned}
        \\
        \added{\protect\ensuremath{(O, \gls{s:ltable})\ \gls{s:satis}}}\ & \added{\protect\ensuremath{\gls{s:s4m+form}_{1} \leq \gls{s:s4m+form}_{2}}}\quad & \added{\protect\ensuremath{\textnormal{iff}}}\quad & \added{\protect\ensuremath{\valuation{\gls{s:s4m+form}_{1}}(O,\gls{s:ltable}) \leq \valuation{\gls{s:s4m+form}_{2}}(O,\gls{s:ltable})}}
        \\
        \added{\protect\ensuremath{(O, \gls{s:ltable})\ \gls{s:satis}}}\ & 
        \added{\protect\ensuremath{\exists{v}(\gls{s:annot}) .\: \gls{s:s4u+form}}}\quad & 
        \added{\protect\ensuremath{\textnormal{iff}}}\quad &
        \begin{aligned}[t]
            & \added[id=gf]{\protect\ensuremath{\exists{o \in \gls{s:objfn}(O, \gls{s:annot}).}}}\\
            & \added[id=gf]{\protect\ensuremath{(O,\gls{s:ltable}(v)\coloneqq o) \: \gls{s:satis}\: \gls{s:s4u+form}}}
        \end{aligned}
    \end{alignat*}
\end{definition}

Notice that the models (i.e., sets of objects) that satisfy a spatial formula \gls{s:s4u+form} are not minimal.
For instance, using the symbols from \cref{ex:spre:semantics:1}, we have $\gls{s:symbol}_{1}\: \gls{s:satis}\: \textnormal{\textit{bus}}$, but also $\gls{s:symbol}_{2}\: \gls{s:satis}\: \textnormal{\textit{bus}}$ and $\gls{s:symbol}_{3}\: \gls{s:satis}\: \textnormal{\textit{bus}}$.
In the \gls{k:pattern-matching} problem, we typically care more so about the sequence of frames $\gls{s:frame}_{i,j}$ that satisfy the pattern \gls{s:query} rather than which exact objects are part of the pattern.

\begin{definition}[\glsxtrshort{a:spre} Semantics]
    \label{def:spre:semantics:3}
    Given the alphabet \gls{s:alphabet}, the language described by a \gls{a:spre} query \gls{s:query} is inductively defined as follows:
    \begin{alignat*}{1}
        \gls{s:langof}(\gls{s:s4u+form}) &= \lbrace \gls{s:symbol} \in \gls{s:alphabet} \mid (\gls{s:symbol},\gls{s:ltable}_{\gls{s:s4u+form}})\: \gls{s:satis}\: \gls{s:s4u+form} \rbrace\\
        \gls{s:langof}(\gls{s:query}_{1}\: \gls{s:alt}\: \gls{s:query}_{2}) &= \gls{s:langof}(\gls{s:query}_{1})\: \gls{s:sunion}\: \gls{s:langof}(\gls{s:query}_{2})\\
        \gls{s:langof}(\gls{s:query}_{1}\: \gls{s:concat}\: \gls{s:query}_{2}) &= \gls{s:langof}(\gls{s:query}_{1})\gls{s:langof}(\gls{s:query}_{2})\\
        \gls{s:langof}(\gls{s:query}{\gls{s:kleene}}) &= \bigcup\nolimits_{i = 0}^{\gls{s:infty}}{\gls{s:langof}(\gls{s:query}^{i})}
    \end{alignat*}
    where $\gls{s:query}^{i}$ denotes the concatenation of pattern \gls{s:query} a total of $i$ times, \added[id=gf]{and $\gls{s:ltable}_{\gls{s:s4u+form}}$ is a look up table initialized to $\bot$ for all variables appearing in $\gls{s:s4u+form}$}.
\end{definition}

One notable difference from the standard language definition for a RE is that now our base case, i.e., the spatial formulas \gls{s:s4u+form} evaluate to sets of symbols as demonstrated in \cref{ex:spre:semantics:1}.
This reflects the observation that at each frame we may have several matching objects for our query.


%% file: src/sections/matching.tex
\section{Perception Stream Matching}
\label{sec:matching}

In this section, we provide a formulation of the problem of pattern matching against perception streams in both the \gls{k:offline} and \gls{k:online} domain. Furthermore, we introduce the \gls{a:strem} tool as our matching framework that implements the semantics of \glspl{a:spre} introduced in \cref{sec:spre} to search over perception streams.

\subsection{Problem Formulation}
\label{sec:matching:problem}

Informally, the traditional problem of \gls{k:pattern-matching} considers a finite word \gls{s:word} and a pattern \gls{s:pattern} from some finite alphabet \gls{s:alphabet} such that the goal is to find all non-overlapping subsets of \gls{s:word} that contain an exact match of \gls{s:pattern}.
From \gls{a:bm} \cite{boyer1977fast} to \gls{a:kmp} \cite{knuth1977fast}, many algorithms have been developed to solve this problem of searching through strings \cite{alfred2014algorithms}.
In this work, we extend upon this idea with the modification that our search pattern \gls{s:pattern} is symbolically a \gls{a:spre} query \gls{s:query}, and our word \gls{s:word} is a perception stream \gls{s:datastream}.
We consider this problem in the both \gls{k:offline} and \gls{k:online} domain.

\subsubsection{Offline Matching}
\label{sec:matching:problem:offline}
We consider pattern matching in the \gls{k:offline} domain primarily motivated by the presence of publicly available \gls{a:av}-based perception datasets \cite{sun2020scalability,caesar2020nuscenes,kesten2019woven,pitropov2021canadian,yu2020bdd100k}.
These datasets provide a large suite of perception data collected for and used by \gls{a:ads} applications.
However, to our knowledge, the capabilities and frameworks to search through these datasets for said applications are not well-supported or require a significant effort to do so.
 The \gls{k:offline} \gls{k:pattern-matching} problem for perception streams is formalized below in \cref{prob:matching:problem:offline:1}.

\begin{problem}[Offline Perception Stream Matching]
  \label{prob:matching:problem:offline:1}
  Given a \textit{finite} perception stream \gls{s:datastream} and a \gls{a:spre} query \gls{s:query}, 
  then starting from frame 0, 
  find the set of all non-overlapping leftmost longest frame subsequences $\lbrace\gls{s:frame}_{i_{1},j_{1}}, \ldots, \gls{s:frame}_{i_{n},j_{n}}\rbrace$ in \gls{s:datastream} such that $i_{k} \leq j_{k}$, $j_{k} \leq i_{k+1}$ and $j_{n} \leq \cardinality{\gls{s:datastream}}$ and $\gls{s:frame}_{i_{k},j_{k}} \in \gls{s:langof}(Q)$ for all $k \leq n$.
\end{problem}

\subsubsection{Online Matching}
\label{sec:matching:problem:online}

We also consider pattern matching in the online domain to perform filtering and querying of perception streams generated in realtime.
Applications of such use cases include \gls{k:monitoring} of \glspl{a:av} deployed, \gls{a:cctv} camera alerts, and any perception-based systems generating data where detection of scenarios in realtime are of importance.

Regarding the procedure of matching online, the framework is re-run at every time instance $l$ when a new frame $\gls{s:frame}_l$ is received and returns the maximal query matched up to that point $\gls{s:frame}_{i_{k},j_{k}}$ with $j_k = l$ (or none if no match).
The \gls{k:online} \gls{k:pattern-matching} problem for perception streams is formalized below in \cref{prob:matching:problem:online:1}.

\begin{problem}[Online Perception Stream Matching]
  \label{prob:matching:problem:online:1}
  Given a perception stream \gls{s:datastream} and a \gls{a:spre} query \gls{s:query}, 
  then at every incoming frame $\gls{s:frame}_j$ of \gls{s:datastream},
  find the longest subsequence of frames $\gls{s:frame}_{i,j}$ such that $0 \leq i \leq j$ and $\gls{s:frame}_{i,j} \in \gls{s:langof}(Q)$.
\end{problem}

\subsection{\glsfmtlong{a:strem}}
\label{sec:matching:strem}

Our \gls{a:spre} matching framework follows the same principles as the classic \gls{a:re} matching frameworks \cite{aho2020compilers}.
Standard string matching approaches translate an \gls{a:re} to a \gls{a:dfa} \gls{s:dfa} which is then used to process the strings.
Our framework deviates from the established approaches using \glspl{a:dfa} since each frame contains multiple objects which may satisfy different \gls{s:s4u+} formulas and all potential matches need to be tracked simultaneously.

\begin{example}
Consider the data stream in Table \ref{tab:prelims:stream:1} and assume that we only care about the classes and properties, e.g., we want to find two frames where a \textit{red bus} appears.
If we treat each object in each frame as a symbol of the form {\tt (class, property)}, then the data stream represents $2 \times 3 \times 2 = 12$ strings.
Two example strings from Table \ref{tab:prelims:stream:1} are {\tt (bus,red)(bus,red)(bus,red)} and {\tt (bus,red)(car, sedan)(bus,red)}.
As the length of the data stream increases, the number of strings that we need to consider increases exponentially in the worst case.
\end{example}

Especially in the case of online query matching, an approach that extracts strings from a perception stream to match against a \gls{a:dfa} quickly becomes unmanageable.
In this work, we take a more pragmatic approach which in practice works well.
We treat each syntactically equivalent \gls{s:s4u+} formula as a unique symbol and translate the \gls{a:spre} into an \gls{a:re}.
Even though we can now use the standard \gls{a:re} to \gls{a:dfa} algorithms, the resulting automaton in execution becomes nondeterministic.
This process can be easily visualized through \cref{ex:matching:strem:from} below.

\begin{example}
    \label{ex:matching:strem:from}
    In the following, we use the convention that an atomic proposition (i.e., a set of attributes) is represented as an augmented \gls{k:charclass} familiar to \gls{k:grep}.
    For readability, \gls{s:s4u+} formulas are also surrounded by brackets.
    Using this notation, the formula \texttt{[<nonempty>([:car:]\&[:ped:])]} is an \gls{s:s4u+} formula that is only true when a frame contains a \textit{car} and a \textit{pedestrian} with interesecting bounding boxes.
    Since the operator {\tt <nonempty>} applies only to spatial terms, we know that \texttt{[:car:]\&[:ped:]} is an \gls{s:s4+} subformula where \texttt{\&} is the operator for set intersection.
    
    From the previously introduced notation, we provide the following \gls{a:spre} pattern written below
    \begin{center}
        \texttt{[<nonempty>([:car:] \& [:ped:])]* ([[:truck:]] | [[:car:]]) [[:car:] \& [:bus:]] [[:bus:]]}
    \end{center}
    that matches zero or more (\gls{s:kleene}) frames where a \textit{car} and \textit{pedestrian} overlap, followed by (\gls{s:concat}) a frame with either (\:\gls{s:alt}\:) a \textit{truck} or a \textit{car}, followed by (\gls{s:concat}) one frame that contains a \textit{car} and a \textit{bus}, and ending with (\gls{s:concat}) a frame with a \textit{bus}.
    
    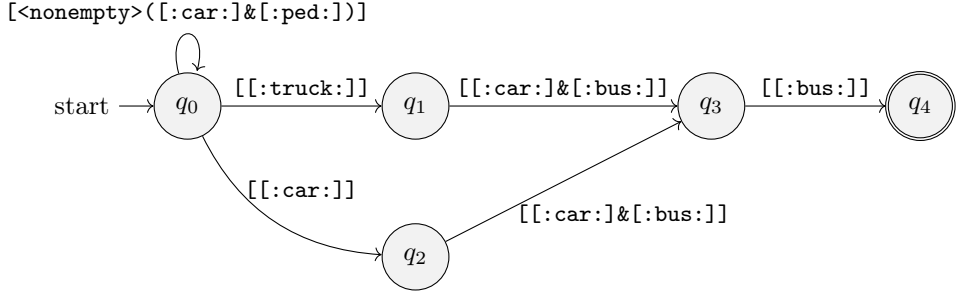
\begin{figure*}[ht]
    \centering
    \begin{tikzpicture}[->,>=To,node distance=3cm,every state/.style={fill=gray!10},initial text=start]
      \node[state, initial] (q1) {$q_{0}$};
      \node[state, right of=q1] (q2) {$q_{1}$};
      \node[state, below = 2cm of q1, right of=q1] (q3) {$q_{2}$};
      \node[state, right = 3cm of q2] (q4) {$q_{3}$};
      \node[state, accepting, right of=q4, node distance=2.75cm] (q5) {$q_{4}$};
      
      \draw (q1) edge[loop above] node{\scriptsize\texttt{[<nonempty>([:car:]\&[:ped:])]}} (q1)
      (q1) edge[above] node{\scriptsize\texttt{[[:truck:]]}} (q2)
      (q1) edge[bend right, above] node[xshift=1em,yshift=0.5em]{\scriptsize\texttt{[[:car:]]}} (q3)
      (q2) edge[above] node{\scriptsize\texttt{[[:car:]\&[:bus:]]}} (q4)
      (q3) edge[below] node[xshift=2.25em,yshift=-0.5em]{\scriptsize\texttt{[[:car:]\&[:bus:]]}} (q4)
      (q4) edge[above] node{\scriptsize\texttt{[[:bus:]]}} (q5);
    \end{tikzpicture}
    \caption{\glsfmttext{a:spre} to \glsfmttext{a:dfa}.}
    \label{fig:matching:strem:from:1}
    \end{figure*}

    The resulting automaton that accepts perception streams that match the  \gls{a:spre} above is presented in \cref{fig:matching:strem:from:1}.
    Notice that in the resulting automaton, the transitions between states are labeled by \gls{s:s4u+} formulas and, hence, the execution semantics is that of a \gls{a:nfa}.
    That is, in each state, multiple transitions may be activated.
    For example, in state $q_0$ if the current frame contains a \textit{truck} and a \textit{car}, then both transitions to $q_1$ and to $q_2$ are activated.
    In principle, tracking multiple states for an \gls{a:nfa} execution scales better than constructing single-symbol strings from a stream for tracking with a \gls{a:dfa}.
    In the worst case, the total number of states of the DFA that we need to keep track off is order of magnitudes smaller than the number of all possible single symbol strings that we need to consider.
\end{example}

\subsubsection{Software Tool}
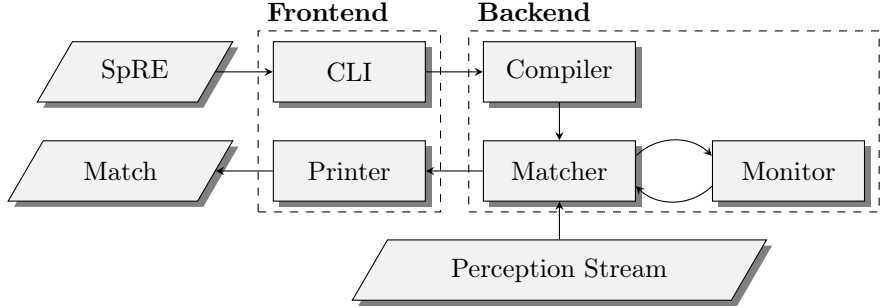
\begin{figure*}[htbp]
  \centering
  \begin{tikzpicture}[node distance=0.5cm and 0.75cm]
    \node[draw,fill=gray!10,drop shadow=black,trapezium,trapezium left angle=60, trapezium right angle=120,minimum width=2cm,minimum height=0.8cm] (n0) {\glsxtrshort{a:spre}};
    \node[draw,fill=gray!10,drop shadow=black,rectangle,minimum width=2cm,minimum height=0.8cm,right=of n0] (n2) {CLI};
    \node[draw,fill=gray!10,drop shadow=black,rectangle,minimum width=2cm,minimum height=0.8cm,below=of n2] (n3) {Printer};
    \node[draw,fill=gray!10,drop shadow=black,trapezium,trapezium left angle=60, trapezium right angle=120,minimum height=0.8cm,left=of n3] (n1) {Match};
    \node[draw,fill=gray!10,drop shadow=black,rectangle,minimum width=2cm,minimum height=0.8cm,right=of n2] (n4) {Compiler};
    \node[draw,fill=gray!10,drop shadow=black,rectangle,minimum width=2cm,minimum height=0.8cm,below=of n4] (n5) {Matcher};
    \node[draw,fill=gray!10,drop shadow=black,rectangle,minimum width=2cm,minimum height=0.8cm,node distance=1cm,right=of n5] (n6) {Monitor};
    \node[draw,fill=gray!10,drop shadow=black,trapezium,trapezium left angle=60, trapezium right angle=120,minimum height=0.8cm,below=of n5] (n7) {Perception Stream};

    \node[draw,rectangle,minimum width=2.4cm,minimum height=2.4cm,dashed] (frontend) at ($(n2)!0.5!(n3)$) {};
    \node[anchor=south west] at (frontend.north west) {\textbf{Frontend}};
    
    \node[draw,rectangle,minimum width=5.4cm,minimum height=2.4cm,dashed] (backend) at ($(n4)!0.5!(n6)$) {};
    \node[anchor=south west] at (backend.north west) {\textbf{Backend}};

    \draw [black,-stealth](n0.east) -- (n2.west);
    \draw [black,-stealth](n2.east) -- (n4.west);
    \draw [black,-stealth](n4.south) -- (n5.north);
    \draw [black,-stealth](n5.west) -- (n3.east);
    \draw [black,-stealth](n3.west) -- (n1.east);
    \draw [black,-stealth](n7.north) -- (n5.south);
    \path [-stealth] 
    ([shift={(0,0.2)}]n5.east) edge[out=45, in=135] ([shift={(0,0.2)}]n6.west)
    ([shift={(0,-0.2)}]n6.west) edge[out=225, in=-45] ([shift={(0,-0.2)}]n5.east);;
  \end{tikzpicture}
  \caption{The architectural design of \glsfmttext{a:strem}.}
  \label{fig:matching:strem:architecture:1}
\end{figure*}

\gls{a:strem} is a \gls{a:cli} tool\footnote{\url{https://crates.io/crates/strem}} 
developed with \gls{k:rust} \cite{matsakis2014rust} to find scenarios of interest in perception streams that match a given \gls{a:spre} query.
It functions in both the \gls{k:offline} and \gls{k:online} domain to search over perception-based datasets or realtime streams, respectively.
An illustration of its core components is provided in \cref{fig:matching:strem:architecture:1}.
As input, the tool accepts a \gls{a:spre} query and a perception stream.
As output, it incrementally returns the set of matches where each match is a range of frames from the provided perception stream that matched the pattern.
The five constituent components of the tool are grouped into two functionalities: the \gls{k:frontend} and the \gls{k:backend}.
The \gls{k:frontend} handles all input-/output-related activities pertinent to the usability of the tool; and the \gls{k:backend} is concerned only with the core matching framework and procedures.
Henceforth, we focus on the \gls{k:backend} components that support the main contributions of this work: the \gls{k:compiler}, \gls{k:matcher}, and \gls{k:monitor} modules.

The \gls{k:compiler} is responsible for translating a \gls{a:spre} into a \gls{a:s-ast}---an \gls{a:ir} form interpretable by the \gls{k:monitor} and \gls{k:matcher} modules.
The \gls{k:monitor} is responsible for evaluating \gls{s:s4u+} formulas against perception stream frames.
The \gls{k:matcher} is responsible for constructing \glspl{a:dfa} from \glspl{a:s-ast} and running the matching algorithms.

\subsubsection{Spatial Matching Algorithms}
\label{sec:matching:strem:pattern}

The pattern matching procedure involves both the \gls{k:matcher} and the \gls{k:monitor}---as depicted in \cref{fig:matching:strem:architecture:1}.
The \gls{k:matcher} receives from the \gls{k:monitor} which \gls{s:s4u+} formulas were satisfied and, then, it takes the appropriate transitions to the next states.
Recall that in our framework multiple transitions on the \gls{a:dfa} may become active. 
Therefore, the execution semantics of the \gls{a:dfa} in the \gls{k:matcher} are effectively the execution semantics of an \gls{a:nfa}.
Nevertheless, our constructions are syntactically \gls{a:dfa} and, in the following, we will still refer to them as DFA.

The algorithm to match against a perception stream given some \gls{a:dfa} that recognizes a valid \gls{a:spre} query is shown in \cref{alg:matching:strem:pattern:1}.
This algorithm is generalized for both \gls{k:offline} and \gls{k:online} applications and any differences in the assumptions and procedures are highlighted in the sections that follow.

\paragraph{Offline Algorithm}
\label{sec:matching:strem:pattern:offline}

The \gls{k:offline} matching procedure matches over a finite perception stream from frame $\gls{s:frame}_{0}$ up to frame $\gls{s:frame}_{l}$ by utilizing a \gls{k:forward} \gls{a:dfa}.
Notably, the \gls{k:offline} variant assumes that all frames within \gls{s:datastream} are present at the beginning of the execution of the matching algorithm.

\paragraph{Online Algorithm}
\label{sec:matching:strem:pattern:online}

The \gls{k:online} matching procedure matches over a perception stream by utilizing a \gls{k:reverse} \gls{a:dfa}.
For each new frame received, the \gls{k:online} algorithm variant is ran.
We use a \gls{k:reverse} \gls{a:dfa} in the \gls{k:online} problem as matching backwards (i.e., from frame $\gls{s:frame}_{l}$ down to frame $\gls{s:frame}_{0}$) ensures that the matching procedure terminates (in the worst case at frame $\gls{s:frame}_{0}$).
However, in practice, it is recommended that the termination of the match be triggered by some finite horizon (i.e., maximum length) for which the \gls{a:spre} query will match up to.
For certain queries, we can compute the finite length needed to determine if a match is possible. 
The length of an \gls{k:online} \gls{a:spre} query can be computed as follows:

\begin{definition}[\glsfmttext{a:spre} Horizon]
  \label{def:matching:strem:pattern:online:1}
  The horizon \gls{s:horizon} of a \gls{a:spre} query \gls{s:query} is inductively defined as follows:
  \begin{align*}
    \replaced{\protect\ensuremath{\gls{s:horizon}(\gls{s:s4u+form})}}{} &\replaced{\protect\ensuremath{= 1}}{}\\
    \replaced{\protect\ensuremath{\gls{s:horizon}(\gls{s:query}_{1}\:\gls{s:alt}\:\gls{s:query}_{2})}}{} & \replaced{\protect\ensuremath{= \textnormal{max}(\gls{s:horizon}(\gls{s:query}_{1}), \gls{s:horizon}(\gls{s:query}_{2}))}}{}\\
    \replaced{\protect\ensuremath{\gls{s:horizon}(\gls{s:query}_{1}\:\gls{s:concat}\:\gls{s:query}_{2})}}{} &\replaced{\protect\ensuremath{= \gls{s:horizon}(\gls{s:query}_{1}) + \gls{s:horizon}(\gls{s:query}_{2})}}{}\\
    \replaced{\protect\ensuremath{\gls{s:horizon}(\gls{s:query}{\gls{s:kleene}})}}{} &\replaced{\protect\ensuremath{= \gls{s:infty}}}{}
  \end{align*}
  where \gls{s:s4u+form} is a spatial formula. 
\end{definition}
When $\gls{s:horizon}(\gls{s:query})$ is finite, i.e., there is no Kleene-star operator, then we know that the \gls{k:online} algorithm will only need to use up to $\gls{s:horizon}(\gls{s:query})$ frames in the past. 
As an alternative to the Kleene-star operator, in our implementation, we provide the \gls{k:range-op} operator to capture bounded-ness (see Sec. \ref{sec:examples}).
If a Kleene-star operator must be used, then a hard bound on the maximum length should be used to keep the monitoring time predictable, in the worst case.

\begin{algorithm} 
  \caption{\textsc{SpatialMatching}\\
  This algorithm represents the \gls{k:offline} variant. For the \gls{k:online} variant, replace each line with its corresponding comment to the right.}
  \label{alg:matching:strem:pattern:1}
  \KwIn{A perception stream \gls{s:datastream}, an initial frame index $i \in \gls{s:datastream}$.}
  \KwOut{A range $[start, end)$ corresponding to the indices from \gls{s:datastream}.}
  \KwData{A set $A$ of distinct active states from the \gls{a:dfa}.}

  \vspace{1.0\baselineskip}
  $start \gets i$\bcp[r]{$end \gets \cardinality{\gls{s:datastream}}$}
  $end \gets start$\bcp[r]{$start \gets end - 1$}

  \ForEach(\bcp[f]{$\gls{s:datastream} = (\gls{s:datastream})$}){$\gls{s:frame} \in \gls{s:datastream}$}{
    \ForEach{($\gls{s:symbol}_{s}$, $\gls{s:syform}$)}{
      \If{\gls{s:frame} \gls{s:satis} \gls{s:syform}}{
        \textit{symbols}.push($\gls{s:symbol}_{s}$)\\
      }
    }

    \ForEach{$\gls{s:symbol}_{s} \in symbols$}{
      $A$.insert($\delta(\gls{s:symbol}_{s})$)
    }

    \uIf{A \KwContains accepting}{
      $end \gets \gls{s:frame}.index$ \bcp[r]{$start \gets \gls{s:frame}.index$}
    }
    \ElseIf{A \KwAll dead}{
      break
    }
  }

  \Return{(start, end)}
\end{algorithm}

\subsubsection{Complexity}
\label{sec:examples:benchmarks:complexity}

The time complexity depends on the data stream \gls{s:datastream} and the query \gls{s:query}.
Let \cardinality{\gls{s:datastream}} be the total number of frames in the perception stream, 
and \cardinality{O_{i}} be the number of objects in the $i$\textsuperscript{th} frame where $O_{i}$ is the set of objects in the frame $\gls{s:frame}_{i} \in \gls{s:datastream}$. 
The query \gls{s:query} is translated into a \gls{a:dfa} \gls{s:dfa} with set of states \gls{s:dfa}$_S$ and transition relation \gls{s:dfa}$_\Delta$ with \cardinality{\gls{s:dfa}_S} denoting the number of states, and \cardinality{\gls{s:dfa}_\Delta} denoting the total number of transitions, respectively.
Recall that the transitions in \gls{s:dfa} are labeled by spatial formulas from \gls{s:query}.
That is, the transitions have the form $(s, \gls{s:s4u+form}_k, s') \in$ \gls{s:dfa}$_\Delta$. 
We denote by \cardinality{\gls{s:s4u+form}_{k}} the size of the parse tree (number of nodes) of \gls{s:s4u+form}$_k$ since evaluating \gls{s:s4u+form}$_k$ will require traversing its parse tree.

We begin by evaluating the time complexity of the \gls{k:monitor}, followed by the \gls{k:matcher}, followed by the combination of the two.
For the \gls{k:monitor} to evaluate a spatial formula \gls{s:s4u+form}$_k$ against a frame, 
if the \gls{s:s4u+} formula \gls{s:s4u+form}$_k$ contains no spatial operations, then its evaluation takes linear time in the tree traversal of \gls{s:s4u+form}$_k$ (number of internal nodes $(\cardinality{\gls{s:s4u+form}_{k}}-1)/2$) and linear time in the number of objects \cardinality{O_{i}} in a frame for each leaf (number of leaves $(\cardinality{\gls{s:s4u+form}_{k}}+1)/2$) in order to find objects with specific attributes.
Thus, the complexity of running the monitor is $\gls{s:bigo}(\cardinality{\gls{s:s4u+form}_{k}}\times \cardinality{O_{i}})$.
As a special case, if \gls{s:query} only contains queries  about class labels, then we can use a hash table storing whether an object of some class appears in a frame or not, giving us $\gls{s:bigo}(1)$ evaluation of the leaves of \gls{s:s4u+form}$_k$.

The \gls{k:matcher} keeps track of the active states in the \gls{a:dfa} and, for each state, checks all the formulas in the outgoing transitions by calling the \gls{k:monitor}.
In the worst case, \cardinality{\gls{s:dfa}_S} states will be active, which implies that all the transitions in \gls{s:dfa}$_\Delta$ must be checked.
Thus, there will be $\gls{s:bigo}(\cardinality{\gls{s:dfa}_\Delta})$ calls to the \gls{k:monitor}.
Since the \gls{k:matcher} will be called \cardinality{\gls{s:datastream}} times, the complexity of the \gls{k:offline} algorithm is
\begin{equation*}
    \gls{s:bigo}\big (\cardinality{\gls{s:datastream}} \times \cardinality{\gls{s:dfa}_\Delta}    \times \max_k (\cardinality{\gls{s:s4u+form}_{k}}) \times  \max_i (\cardinality{O_{i}}) \big )
\end{equation*}
whereas the \gls{k:online} algorithm pays this cost for every new frame that appears within the perception stream.

If spatial operations are present in the spatial formulas, then the worst-case time complexity increases.
Spatial terms in \cref{def:spre:semantics:1} evaluate to collections of bounding sets in 2D or 3D.
Therefore, the leaf nodes of \gls{s:s4u+form}$_k$ represent collections of sets, and the internal nodes apply set operations such as union, intersection, complementation, and set difference.
The computational cost of the set operations depends on the set representation (e.g., orthogonal polyhedra, vertex representation, polytopes, zonotopes, etc).
Here, we will not consider the representation of the sets explicitly, and refer the reader to \cite{hekmatnejad2022formalizing}.

\begin{remark}[Best-Case Scenario]
Our querying problem can be reduced to standard regular expression matching when:
    (1) all \gls{s:s4u+} formulas are strictly atomic propositions, (2) the number of object attributes to search over are few in numbers, and (3) the attributes are not quantitative (e.g., no bounding box).
\end{remark}


%% file: src/sections/experiments.tex
\section{Examples and Benchmarks}
\label{sec:examples}

To demonstrate the application of \gls{a:strem}, we provide two use cases of the tool: (A) an \gls{k:offline} example of searching through the \gls{k:dataset} dataset \cite{kesten2019woven} and (B) an \gls{k:online} example of monitoring an \gls{a:av}'s perception system through the \gls{a:carla} simulator \cite{dosovitskiy2017carla} with \gls{a:ros} \cite{quigley2009ros}.
Furthermore, we provide an initial set of performance benchmarks of the tool.
For all queries, we use the \gls{a:strem} implementation-level syntax equivalents  in \cref{tab:examples:1}.

\begin{table}[htbp]
    \centering
    \setlength{\tabcolsep}{0.5pt}
    \caption{\glsfmttext{a:spre} implementation equivalencies}
    \begin{tabular*}{\linewidth}{@{\extracolsep{\fill}}lccccccccc}
        \toprule
        \textbf{Notation} & \gls{s:alt} & \gls{s:concat} & \gls{s:kleene} & \gls{s:isnonempty} & \gls{s:not} & \gls{s:and} & \gls{s:or} & \gls{s:inter} & \gls{s:union}\\
        \midrule
        \textbf{Symbol} & \texttt{|} & & \texttt{*} & \texttt{<nonempty>} & \texttt{\raisebox{0.5ex}{\texttildelow}} & \texttt{\&} & \texttt{|} & \texttt{\&} & \texttt{|}\\
        \bottomrule
    \end{tabular*}
    \label{tab:examples:1}
\end{table}

Furthermore, the \gls{k:range-op} meta-operator $(\{m,n\})$ is used to support constraint concatenations.
The operational equivalence is shown below:

\begin{align*}
    \gls{s:query}\lbrace m, n \rbrace  \equiv& \overbrace{Q\: \gls{s:concat}\: \ldots\: \gls{s:concat}\: Q}^\text{$m$ concatenations} \, | \, 
    \overbrace{Q\: \gls{s:concat}\: \ldots\: \gls{s:concat}\: Q}^\text{$m+1$ concatenations} \, |\\*
    & \ldots \, | \, \overbrace{Q\: \gls{s:concat}\: \ldots\: \gls{s:concat}\: Q}^\text{$n$ concatenations}
\end{align*}
where $0 \leq m \leq n$.
In addition, the \gls{k:range-op} operator support two other functions: (1) \gls{s:query}\{m\} matches \gls{s:query} exactly $m$ times; and (2) \gls{s:query}\{m,\} = \gls{s:query}\{m,$\infty$\}  matches \gls{s:query} $m$ or more times.

\subsection{Example A: Offline Matching Examples}
\label{sec:examples:offline}

We demonstrate the \gls{k:offline} searching capabilities of the \gls{a:strem} tool on the \gls{k:dataset} dataset: a collection of sensor and ground-truth labels used in training and evaluation of \gls{a:av} perception systems.
The dataset is comprised of 10 sensor channels, 360 scenes, 9 object \replaced{classification categories}{classifications}, over 300K \replaced{images}{frames}, and over 1.2M object annotations yielding slightly over 186 GBs of data to search from.
\added{In the examples that follow, we showcase results that come from a single channel of the perception system from a matching frame.}

\subsubsection{Example A.1}
\label{sec:examples:offline:a:1}

In \gls{a:ads} applications, it is important to distinguish between cyclists and pedestrians as the intent and behavior of both differ.
Therefore, to improve the resilience of a perception system against mis-identification, filtering for scenarios where the two classifications overlap (i.e., potential cases of ambiguity) \replaced{assist in strengthening}{strengthens} \glspl{a:dnn} \added{classification behaviors} on such edge cases.

\begin{query}
  \label{query:examples:offline:1}
  Find all \added{instances of the longest consecutive} \replaced{sequence}{sequences} of frames where a \deleted{detected} \textit{pedestrian} overlaps with a \deleted{detected} \textit{cyclist}.

  \begin{center}
    \texttt{[<nonempty>([:pedestrian:] \&}
    \texttt{[:bicycle:])]*}
  \end{center}
  where the \gls{a:spre} matches zero or more frames (\replaced{\texttt{*}}{\gls{s:kleene-start}}) where the intersection \added{(\texttt{\&})} of a \textit{pedestrian} and \textit{bicycle} bounding box is non-empty \added{(\texttt{<nonempty>})}.
\end{query}

\begin{figure}[htbp]
  \centering
  \begin{minipage}[c]{0.49\linewidth}
    \includegraphics[width=1.0\linewidth]{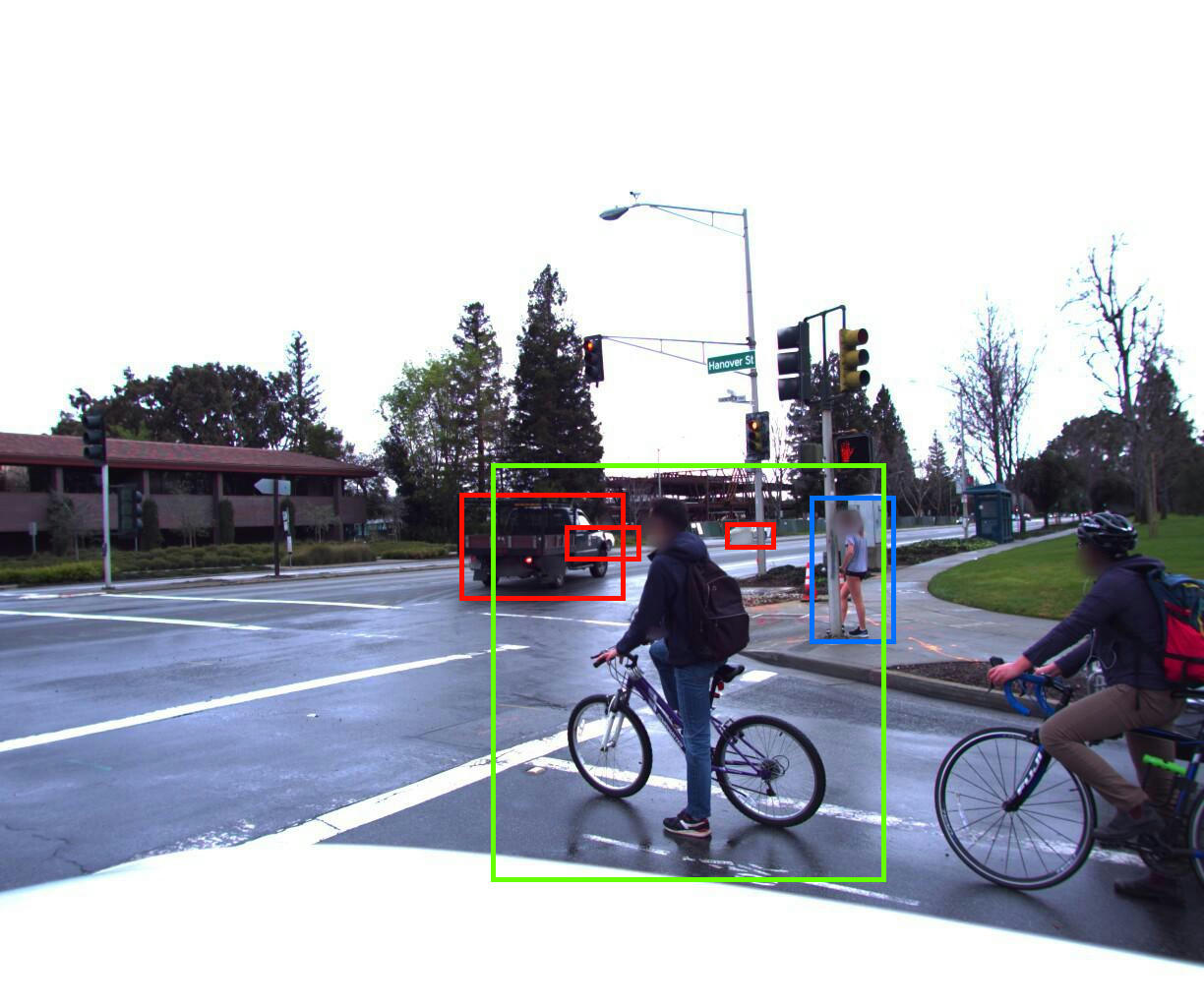}
  \end{minipage}
  \begin{minipage}[c]{0.49\linewidth}
    \includegraphics[width=1.0\linewidth]{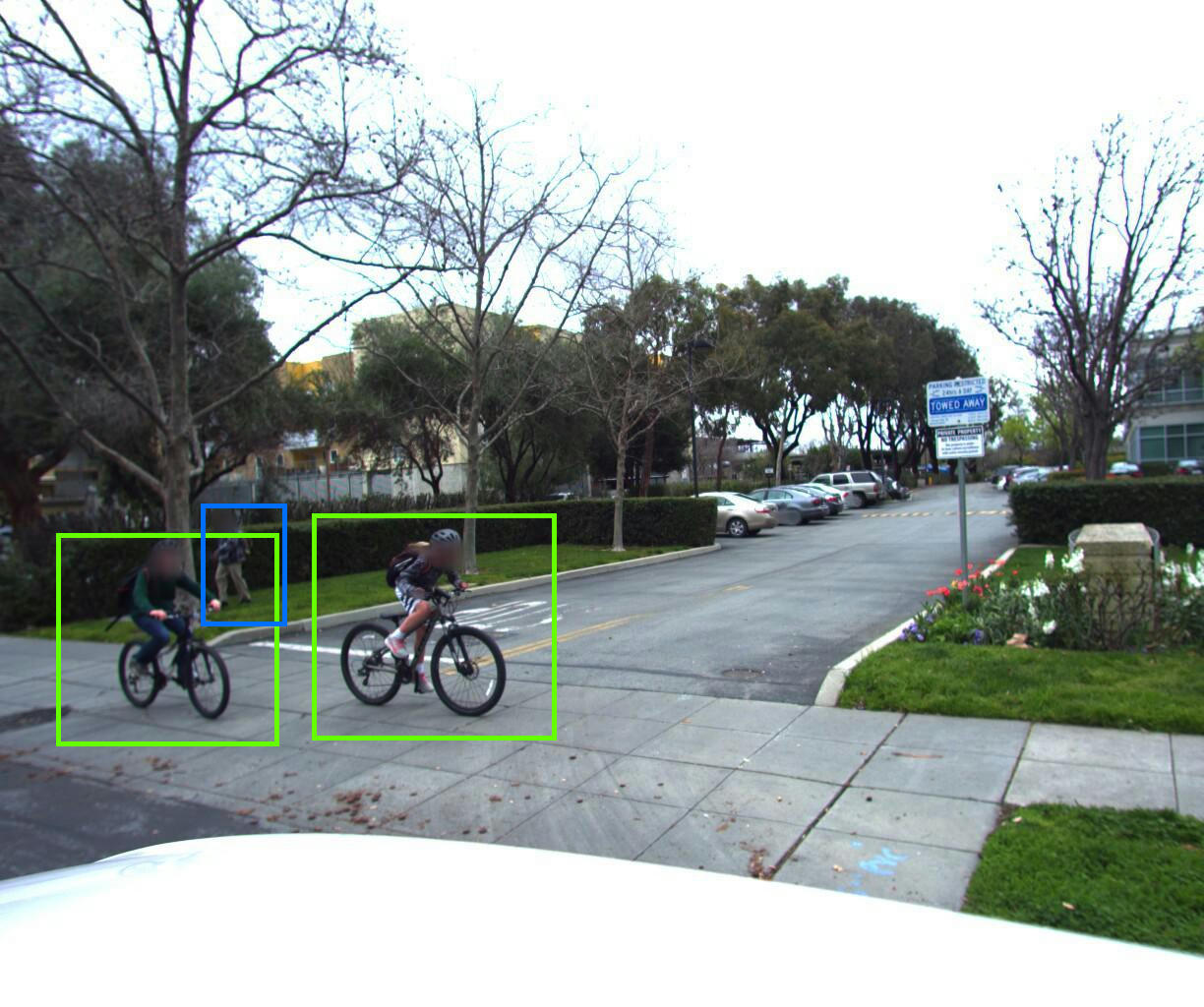}
  \end{minipage}
  \begin{minipage}[c]{1.0\linewidth}
    \includegraphics[width=1.0\linewidth]{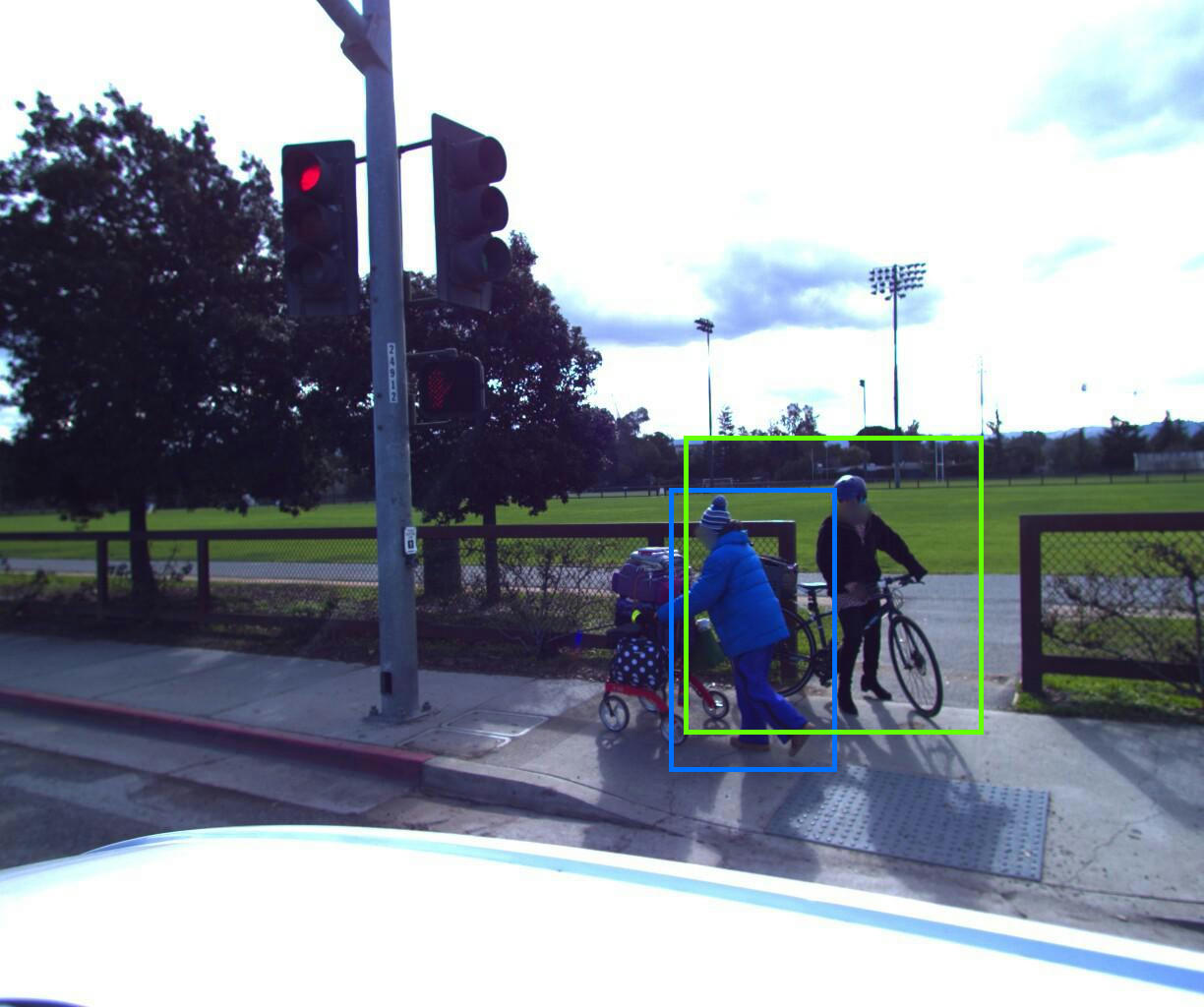}
  \end{minipage}\hfill
  \caption{A selection of three separately matching frames from the \glsfmttext{k:dataset} dataset where a \textit{pedestrian} (blue) overlaps with a \textit{cyclist} (green).}
  \label{fig:examples:offline:1}
\end{figure}

From the results, a total of 62 unique matches were found where each match contains a sequence of frames with a pedestrian and cyclist overlapping \added{from a single channel}.

\subsubsection{Example A.2}
\label{sec:examples:offline:a:2}

While queries targeting individual scenarios are useful for simple matching, more complex queries are needed to capture scenarios that \replaced{cannot}{can not} be consolidated to a single frame and represent \replaced{a series}{an evolution} of \added{different} events.
For instance, consider the scenario where \replaced{some}{a} pedestrian is initially occluded by \replaced{some}{a} \replaced{car}{vehicle}, \added{followed by a frame with} \replaced{some}{an} unobstructed \added{pedestrian}, and then \added{followed by a frame with some pedestrian} occluded again by \replaced{some}{a} \replaced{car}{vehicle}.\footnote{\added{We are deliberate in the use of the word \textit{some} when talking about objects to indicate that a \gls{a:spre} query does not currently support quantification of objects across frames. We recommend \cite{hekmatnejad2022formalizing} for validating the results filtered by \gls{a:strem}, accordingly.}}

\begin{query}
  \label{query:examples:offline:2}
  Find a sequence of frames where a \textit{pedestrian} and \textit{car} occlusion occurs for one or more frames, followed by an unobstructed \textit{pedestrian} for one or more frames, followed by an occlusion of a \textit{pedestrian} and a \textit{car} for one or more frames.
    
  \begin{center}
    \texttt{[<nonempty>([:pedestrian:] \& [:car:])]\{1,\} [[:pedestrian:] \& \raisebox{0.5ex}{\texttildelow}<nonempty>([:pedestrian:] \& [:car:])]\{1,\}[<nonempty>([:pedestrian:] \& [:car:])]\{1,\}}
  \end{center}
  where the \gls{a:spre} matches a sequence of scenarios (i.e., sub-scenario) where each sub-scenario must be at least one frame (\replaced{\texttt{\{1,\}}}{\{1, \}}) long.
  The first sub-scenario matches the intersection \added{(\texttt{\&})} of a \textit{pedestrian} and \textit{car} bounding box is non-empty \added{(\texttt{<nonempty>})}.
  The second sub-scenario matches the case where a \textit{pedestrian} \replaced{is present}{exists} and the intersection of a \textit{pedestrian} with a \textit{car} is not non-empty (i.e., is empty) \added{(\texttt{\raisebox{0.5ex}{\texttildelow}<nonempty>})}.
  The last sub-scenario matches the same as the first sub-scenario.
\end{query}

\begin{figure}[htbp]
  \centering
  \begin{minipage}[c]{0.48\linewidth}
    \includegraphics[width=1.0\linewidth]{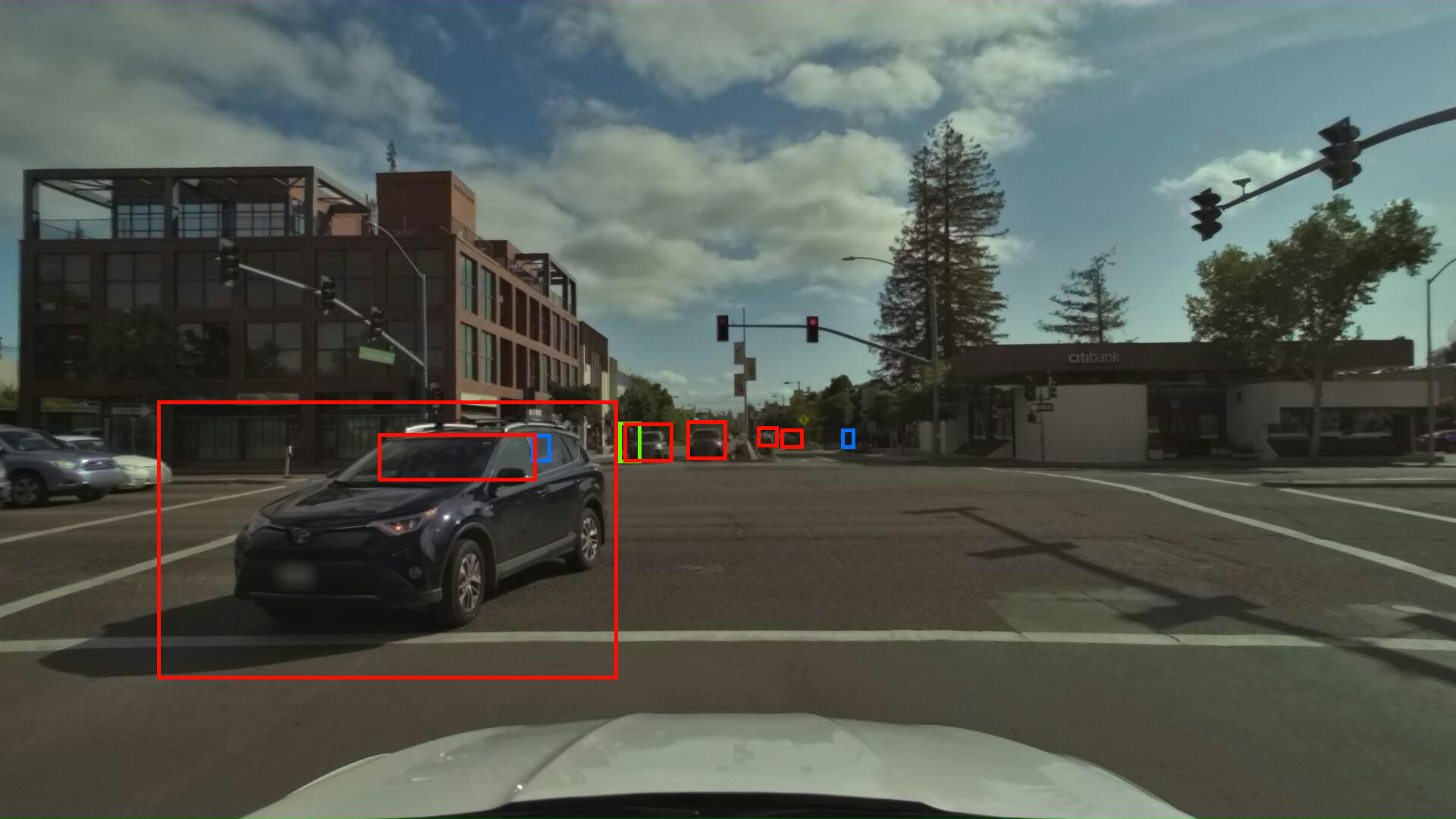}
    \centering
    $\gls{s:frame}_{9}$
  \end{minipage}\hfill
  \begin{minipage}[c]{0.48\linewidth}
    \includegraphics[width=1.0\linewidth]{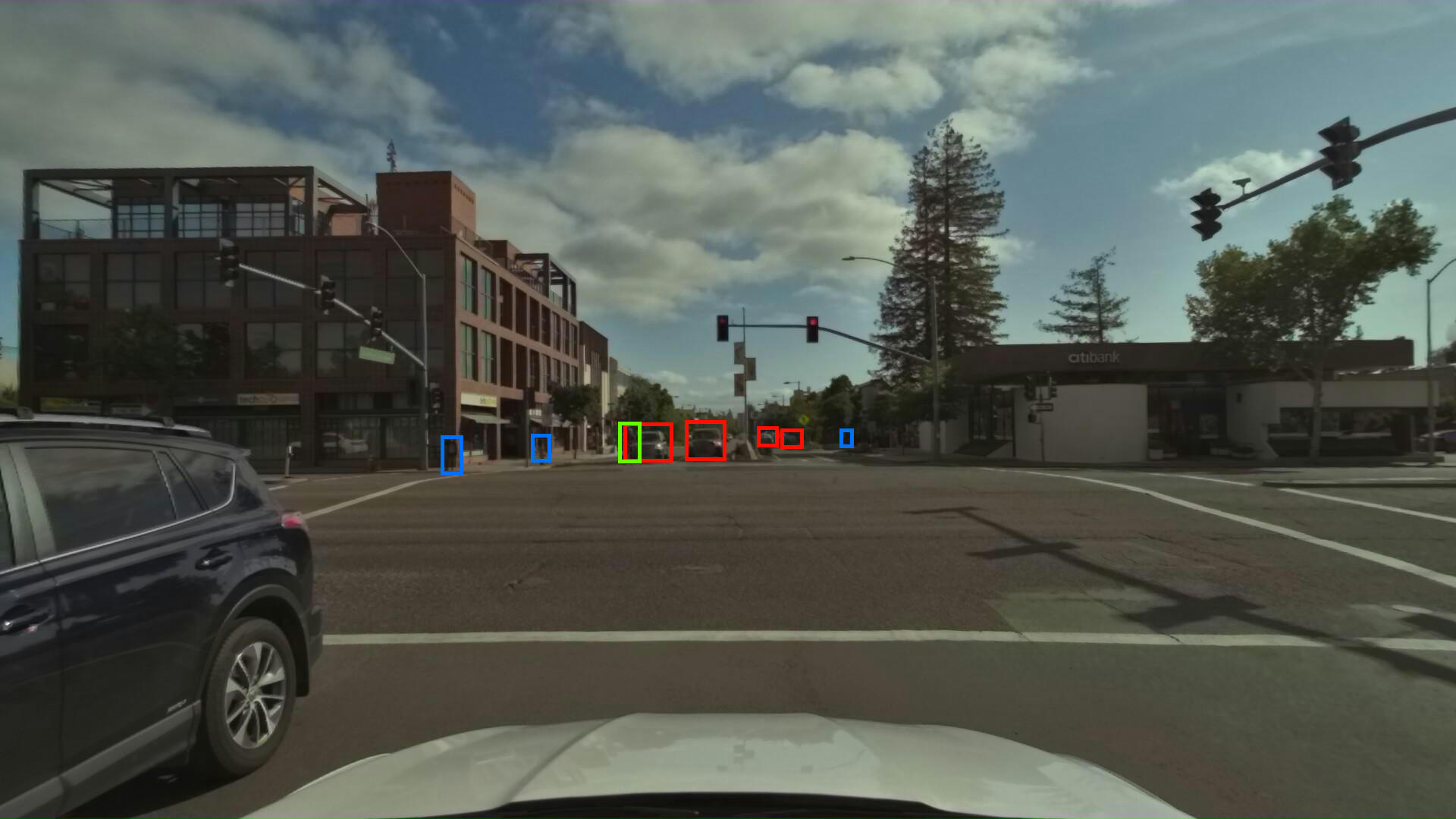}
    \centering
    $\gls{s:frame}_{13}$
  \end{minipage}\newline\vspace{0.5em}
  \begin{minipage}[c]{1.0\linewidth}
    \includegraphics[width=1.0\linewidth]{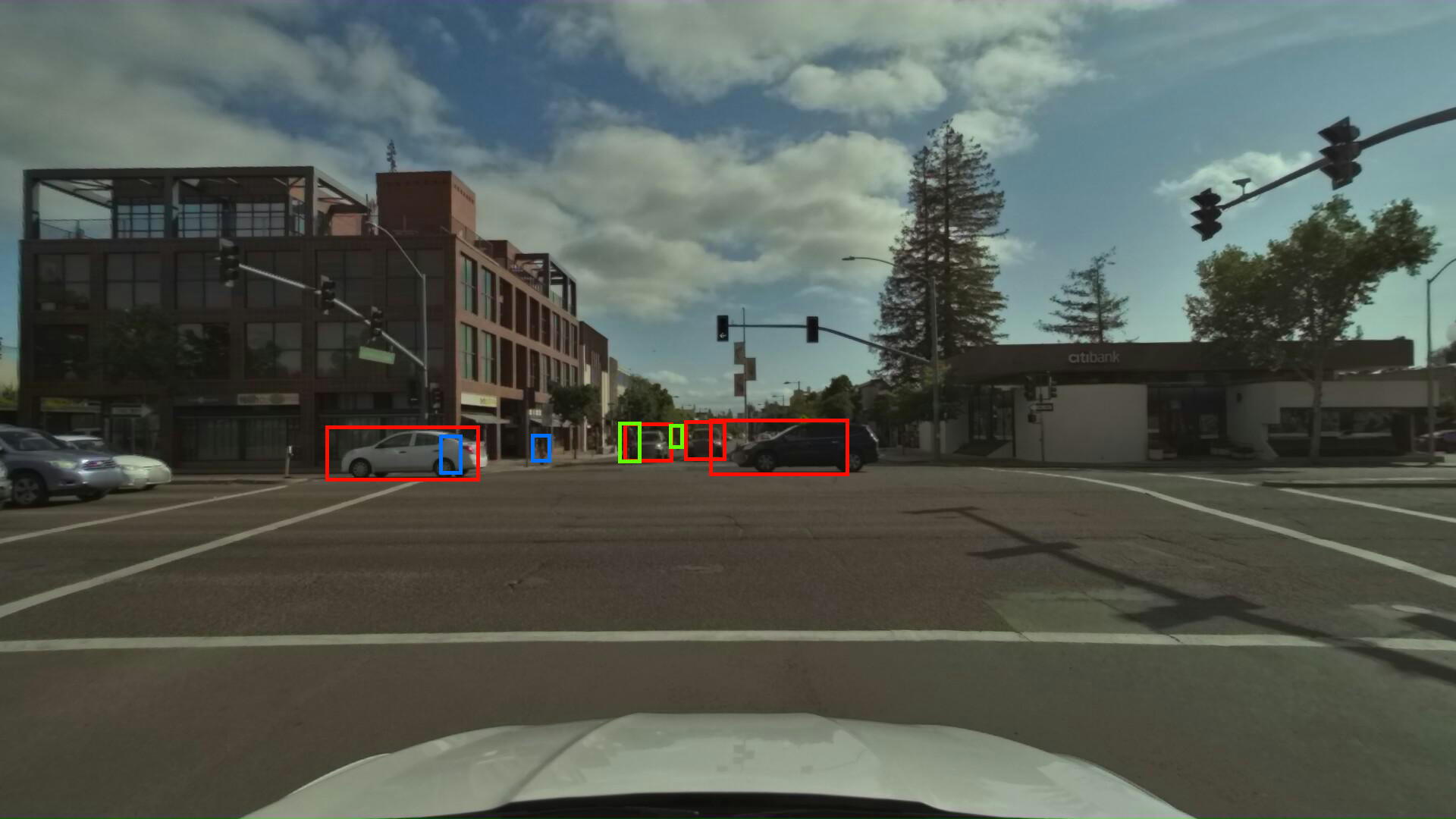}
    \centering
    $\gls{s:frame}_{34}$
  \end{minipage}\vspace{0.5em}
  \caption{The results of running \cref{query:examples:offline:2} through \glsfmttext{a:strem} on the \glsfmttext{k:dataset} dataset. From left to right, the matching frames include an instance of a \textit{car} (red) occluding a \textit{pedestrian} (blue), a \textit{pedestrian}, and a \textit{car} occluding a \textit{pedestrian}.}
  \label{fig:examples:offline:2}
\end{figure}

From the results, a total of 336 uniques matches of three or more frames were found that matched the evolution of scenarios as described in the \gls{a:spre} query.

\subsubsection{Example A.3}
\label{sec:examples:offline:a:3}

\added{As \gls{a:av} datasets include manually annotated information that capture positional-based information of the objects within a scene, the possibilities for searching and filtering for safety-critical scenarios are widened.}
\added{For example, consider the scenario of a pedestrian standing within the road next to a vehicle within some close proximity such that this occurs in front of the \gls{a:av} (i.e., the \gls{a:av} must proceed with extra caution to ensure safety is upheld and avoid careless injury).}

\begin{query}
  \label{query:examples:offline:3}
  \added{Find a sequence of frames where a \textit{pedestrian} is standing close to the left-hand side of a \textit{truck} in front of the \textit{ego} vehicle.}
    
  \begin{center}
    \added{\texttt{[<exists>(p := [:pedestrian:])(<exists>(q := [:truck:])(<y>(p) < <y>(q) \& <dist>(p, q) < 2.0 \& <x>(p) > <x>([:ego:]))))]*}}
  \end{center}
  \added{where the \gls{a:spre} matches zero or more (\texttt{*}) frames where a \textit{pedestrian} is standing to the left-hand side (\texttt{<leftof>}) of a \textit{truck} within two meters (\texttt{<dist>}) such that this occurs in front of (\texttt{<frontof>}) the \textit{ego} vehicle.}
\end{query}

\begin{figure}[htbp]
  \centering
  \begin{minipage}[c]{0.49\linewidth}
    \includegraphics[width=1.0\linewidth]{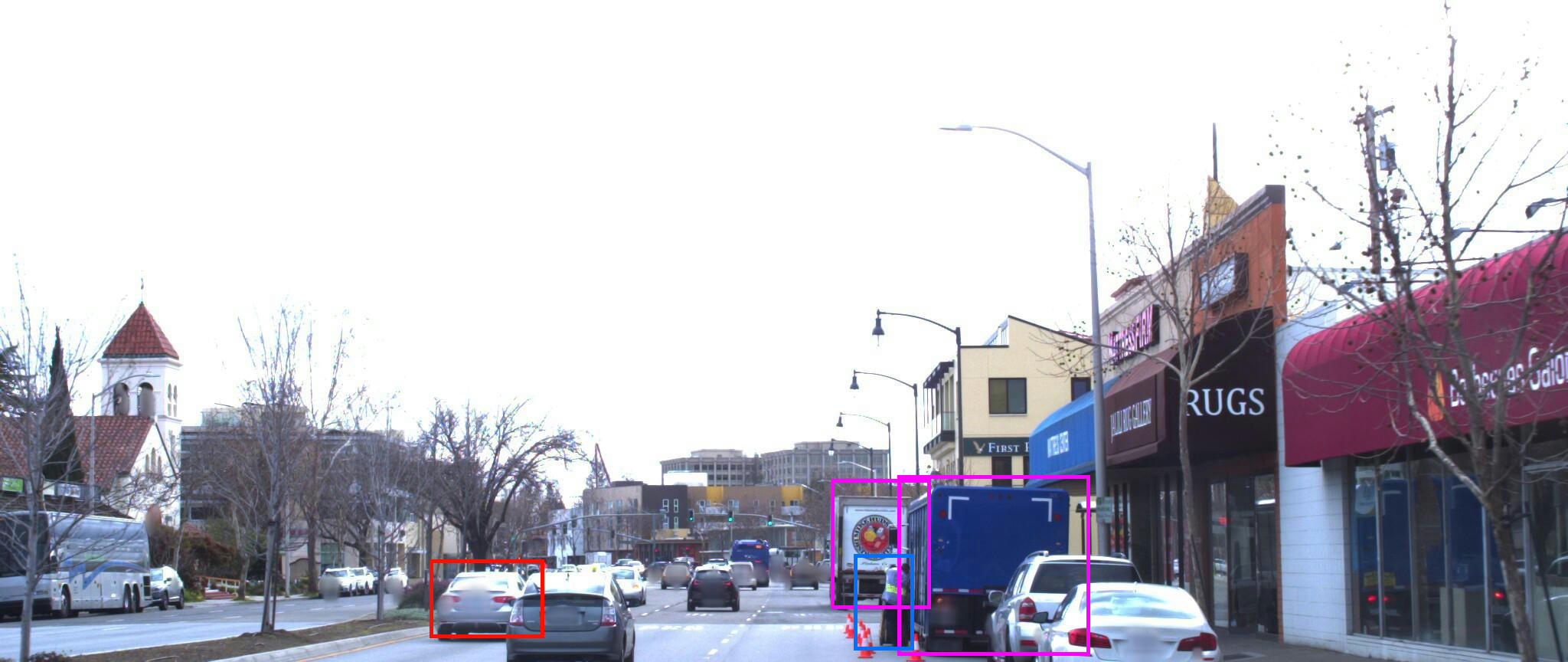}
    \centering
    $\gls{s:frame}_{117}$
  \end{minipage}\hfill
  \begin{minipage}[c]{0.49\linewidth}
    \includegraphics[width=1.0\linewidth]{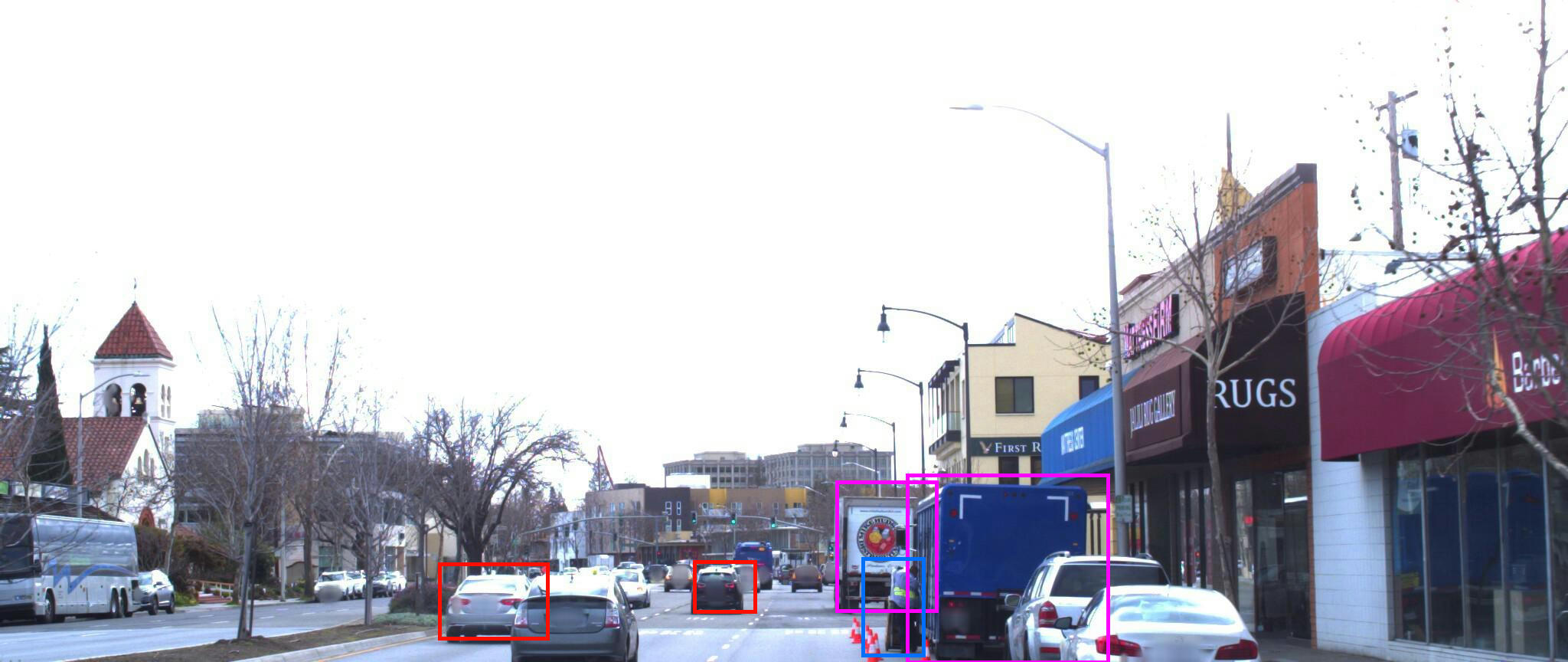}
    \centering
    $\gls{s:frame}_{118}$
  \end{minipage}\newline\vspace{0.5em}
  \begin{minipage}[c]{1.0\linewidth}
    \includegraphics[width=1.0\linewidth]{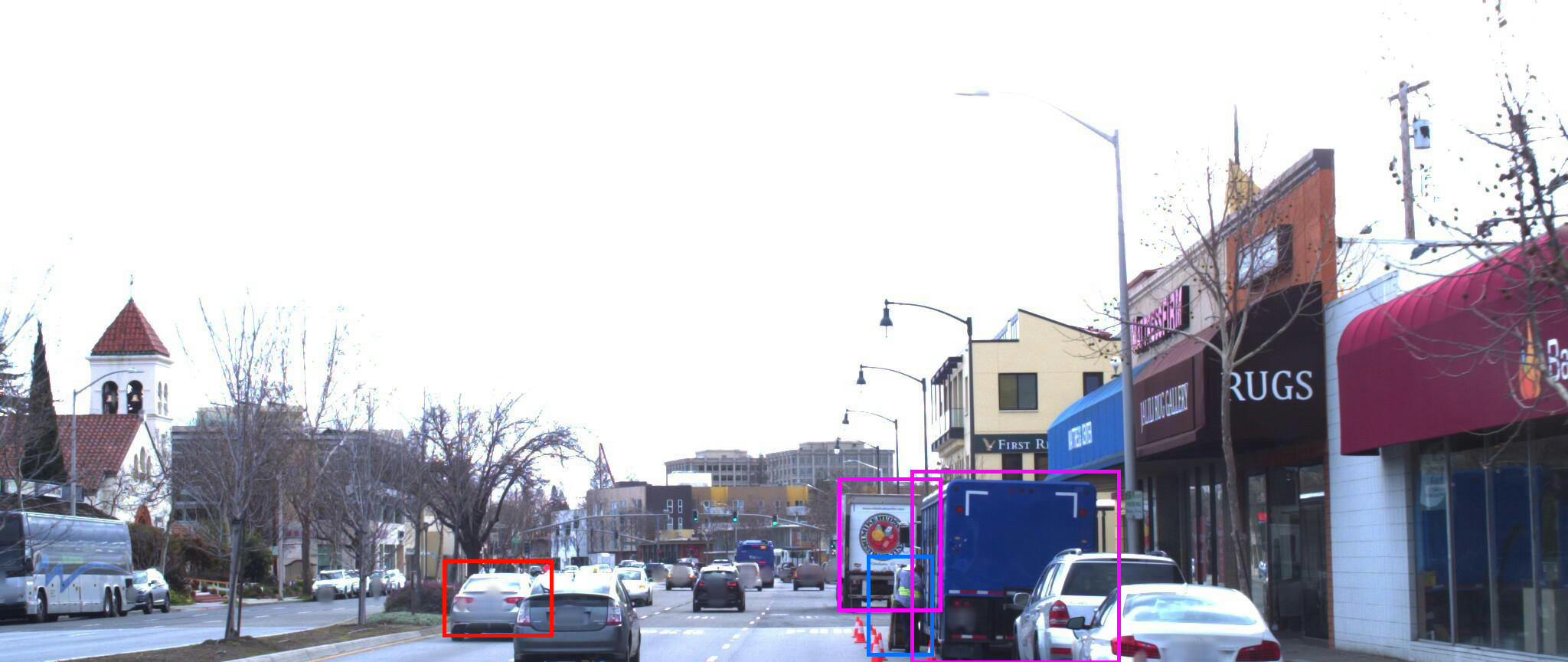}
    \centering
    $\gls{s:frame}_{119}$
  \end{minipage}\vspace{0.5em}
  \caption{The results of running \cref{query:examples:offline:3} through \glsfmttext{a:strem} on the \glsfmttext{k:dataset} dataset. At each frame, a \textit{pedestrian} (blue) is near a \textit{truck} (magenta) in front of the ego vehicle (i.e., front facing camera).}
  \label{fig:examples:offline:3}
\end{figure}

\subsection{Example B: Online Matching}
\label{sec:examples:online}

To demonstrate the online searching capabilities of \gls{a:strem}, we developed a \gls{a:ros} package that bridges the \gls{a:carla} simulator with the \gls{a:strem} tool by using the standard \gls{k:topics} infrastructure provided by \gls{a:ros}.
This design allows additional \gls{a:ros} applications (e.g., robots, \glspl{a:av}, etc.) to easily integrate and subscribe to the match results published by the \gls{a:strem} tool.

\subsubsection{Simulator Setup}

For each example, the \gls{a:carla} server was populated with 50 vehicles (e.g., trucks, sedans, etc.), 20 walkers (i.e., pedestrians), and a single \gls{k:ego} vehicle affixed with a one front-facing camera sensor.
From the set of labels provided by \gls{a:carla}, we capture bounding box information for the following actor types: (1) traffic signs, (2) traffic lights, (3) vehicles, and (4) walkers.

For all examples, the experiments were run on a Linux workstation running Ubuntu 20.04.6 with an AMD Ryzen 7 5800X, an NVIDIA GeForce RTX 3070, and 16 GBs of RAM with \gls{a:carla} v0.9.13 at 60Hz and \gls{a:ros} Noetic (Focal).

\subsubsection{Example B.1}
\label{sec:examples:online:b:1}

Within the deployment of \glspl{a:av}, monitoring perception streams for critical scenarios is a runtime-centric activity that requires a continuous analysis of the results of the perception system in order to take decisive actions quickly.
An example of such a critical scenario common to \glspl{a:av} is in the occlusion of people by other vehicles in the scene.
Within this situation, limited information is available to the system and naturally additional caution should be taken.
However, reporting this information is not an inherent responsibility of the perception system.
As such, the \gls{a:strem} tool provides the capability to instantaneously report frames in realtime where a pedestrian is occluded by some other object detected within a scene to allow the \gls{a:ads} to take action.

In this example, we consider the scenario in \gls{a:carla} where a bounding box of a pedestrian and a vehicle annotation overlap one another.
The formalization of this pattern is presented in \cref{query:examples:online:b:1:1} below.

\begin{query}
  \label{query:examples:online:b:1:1}
  Report every frame where a \textit{pedestrian} and \textit{vehicle} overlap.

  \begin{center}
    \texttt{[<nonempty>([:pedestrian:]\&[:vehicle])]}
  \end{center}
  where the \gls{a:spre} matches a single frame such that the intersection of the bounding boxes of a \textit{pedestrian} and a \textit{vehicle} classification is non-empty.
\end{query}

\begin{figure}[!htbp]
  \centering
    \begin{minipage}[c]{0.49\linewidth}
      \includegraphics[width=1.0\linewidth]{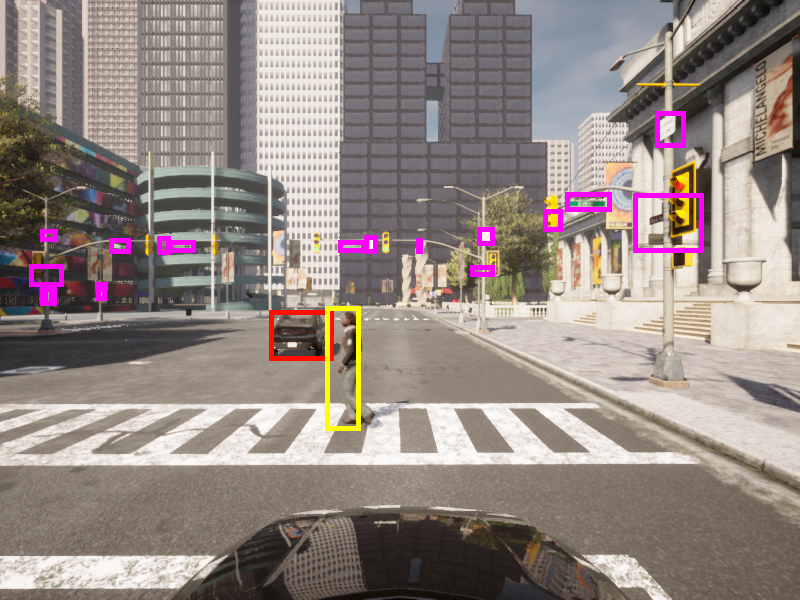}
      \centering
      $\gls{s:frame}_{9354}$
    \end{minipage}\hfill
    \begin{minipage}[c]{0.49\linewidth}
      \includegraphics[width=1.0\linewidth]{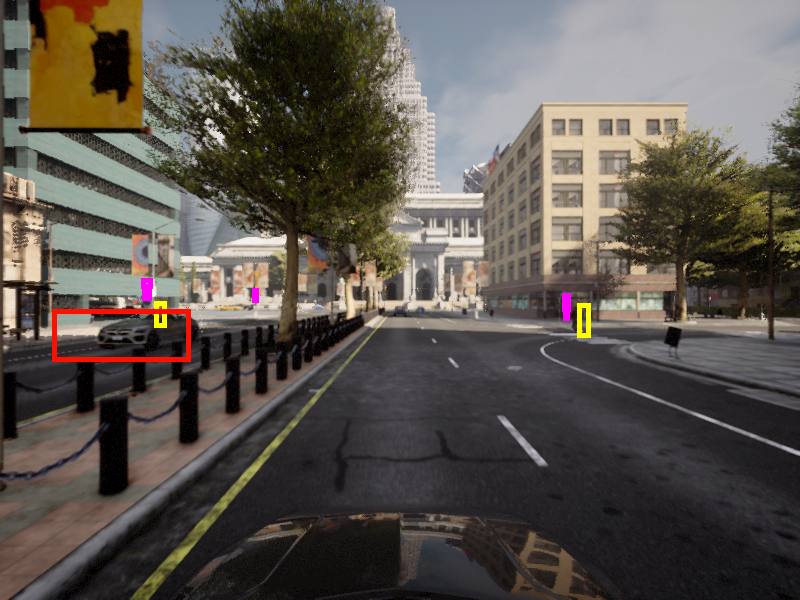}
      \centering
      $\gls{s:frame}_{15407}$
    \end{minipage}\newline\vspace{0.5em}
    \begin{minipage}[c]{1.0\linewidth}
      \includegraphics[width=1.0\linewidth]{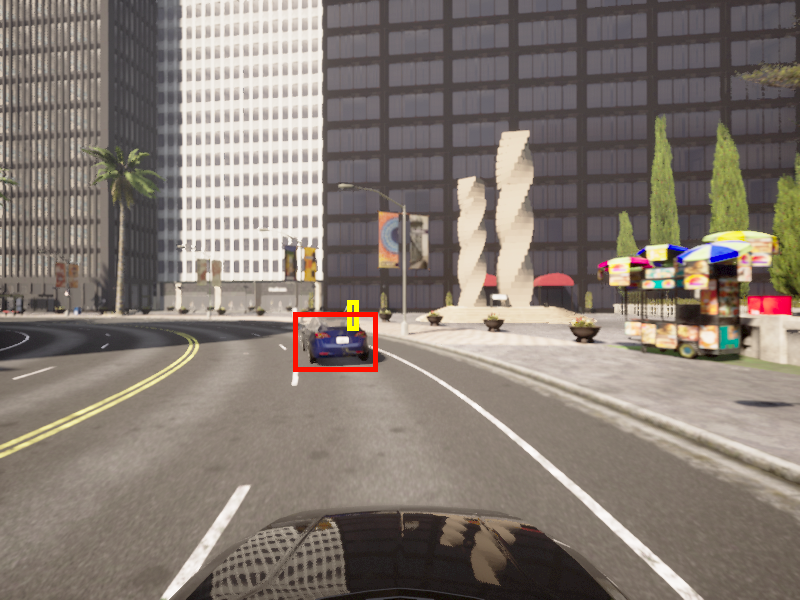}
      \centering
      $\gls{s:frame}_{16869}$
    \end{minipage}\vspace{0.5em}
    \caption{A series of matching frames with object detections within the \glsfmttext{a:carla} simulator where a \textit{pedestrian} and \textit{vehicle} intersect as expressed in \cref{query:examples:online:b:1:1}.}
    \label{fig:examples:online:1}
\end{figure}

A illustrative example of some frames reported by \gls{a:strem} during the simulation are showcased in \cref{fig:examples:online:1}.

\subsubsection{Example B.2}
\label{sec:examples:online:b:2}

Another critical scenario that a perception system may experience is in an eventual case that information becomes missing (i.e., the presence of an object disappears from sight).

In this example, we consider the scenario in \gls{a:carla} where perceived traffic signs are detected followed by an occlusion of some sign within the frame.

\begin{query}
  \label{query:examples:online:b:1:2}
  Find a traffic \textit{sign} within the last 200 frames that is eventually occluded by a \textit{vehicle} or \textit{pedestrian}.
  
  \begin{center}
    \texttt{[[:sign:]]\{1,200\} [<nonempty>(([:vehicle:] | [:pedestrian:]) \& [:sign:])]}
  \end{center}
  where the \gls{a:spre} matches at least 1 and at most 200 frames (\{1, 200\}) initially that contain a \textit{sign} annotation, and ends with one frame where the resulting intersection of the bounding box of a \textit{sign} with the bounding box of either a \textit{vehicle} or \textit{pedestrian} is non-empty.
\end{query}

\begin{figure}[htbp]
  \centering
    \begin{minipage}[c]{0.49\linewidth}
      \includegraphics[width=1.0\linewidth]{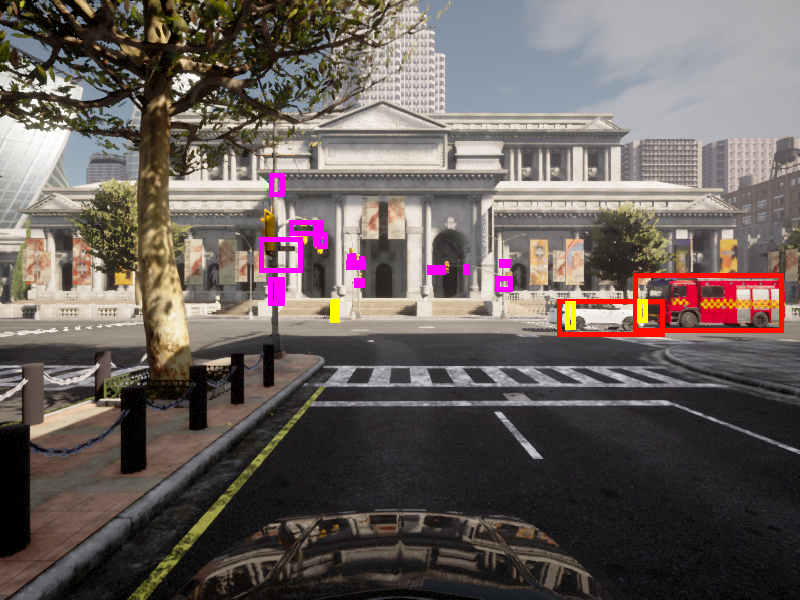}
      \centering
      $\gls{s:frame}_{22215}$
    \end{minipage}\hfill
    \begin{minipage}[c]{0.49\linewidth}
      \includegraphics[width=1.0\linewidth]{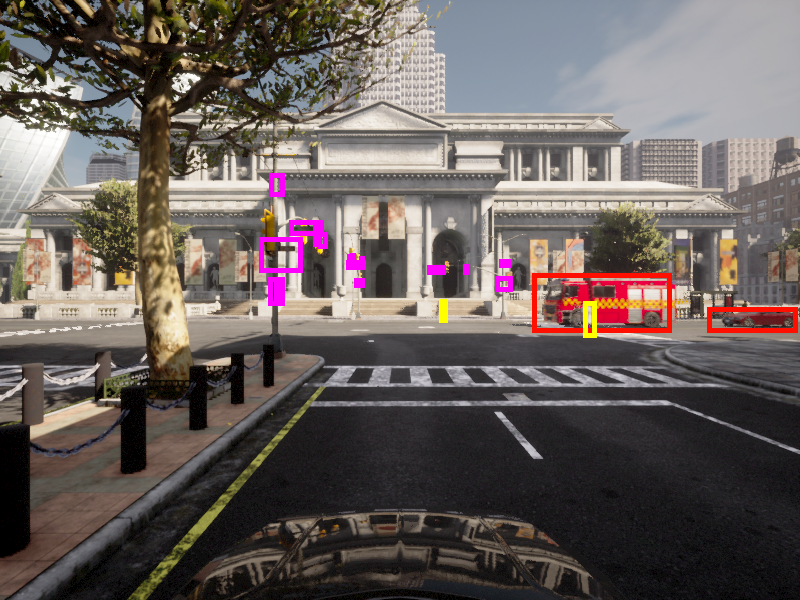}
      \centering
      $\gls{s:frame}_{22341}$
    \end{minipage}\newline\vspace{0.5em}
    \begin{minipage}[c]{1.0\linewidth}
      \includegraphics[width=1.0\linewidth]{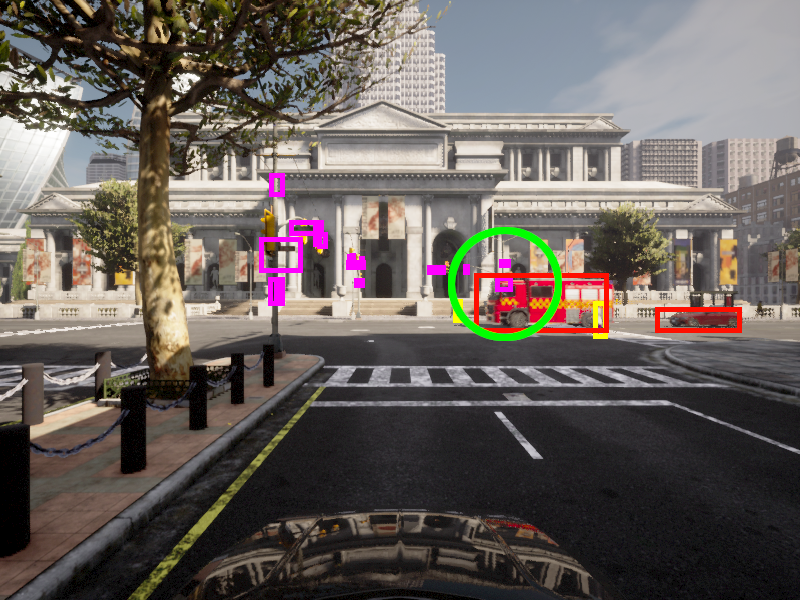}
      \centering
      $\gls{s:frame}_{22358}$
    \end{minipage}\vspace{0.5em}
    \caption{A series of matching frames with object detections within the \glsfmttext{a:carla} simulator where a vehicle (red) eventually occludes a traffic sign (pink) in the green circle.}
    \label{fig:examples:online:2}
\end{figure}

A illustrative example of some frames reported by \gls{a:strem} during the simulation are showcased in \cref{fig:examples:online:2}.

\subsubsection{Example B.3}
\label{sec:examples:online:b:3}

\added{The data generated by a perception stream may not always be confined to an image space as the working environment where all annotations have an \gls{a:aabb}.}
\added{In many cases, real-world applications and deployment of \gls{a:av} and robotic systems rely on a 3D-based perceived environment that is captured from multiple sensor arrays and sensor fusions to contextualize themselves in real-time.}

\added{In this example, we consider the scenario in \gls{a:carla} where an \gls{a:av} (i.e., ego) perceives the world through multiple sensors to re-construct a \gls{k:top-down} overview of its working environment to search for scenarios where a \textit{cyclist} gets too close to the front-left of the ego for five seconds.}

\begin{query}
  \label{query:examples:online:b:1:3}
  \added{Report all frames where a \textit{cyclist} is too close to the front-left side of the \textit{ego} vehicle for 300 frames.}
  
  \begin{center}
    \added{\texttt{[<exists>(v := [:cyclist:])(<x>(v) > <x>([:ego:]) \& <y>(v) < <y>([:ego:]) \& <dist>(v, [:ego:]) < 1.0]\{300\}}}
  \end{center}
  \added{where the \gls{a:spre} matches exactly 300 frames (\texttt{\{300\}}) that contain a \textit{cyclist} spatially oriented in the front, left of the \textit{ego} car within one meter.}
  \added{Without the use of the existential operator, performing multiple operations contextualized to the same object would not be possible.}
\end{query}

\begin{figure}[htbp]
  \centering
    \begin{minipage}[c]{0.49\linewidth}
      \includegraphics[width=1.0\linewidth]{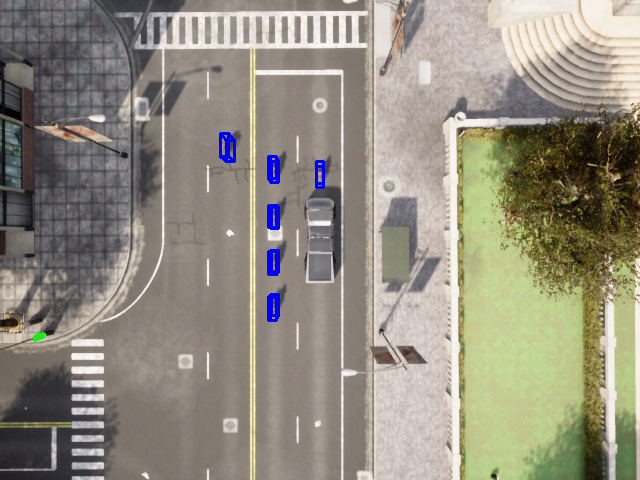}
      \centering $\gls{s:frame}_{4356, 4656}$
    \end{minipage}\hfill
    \begin{minipage}[c]{0.49\linewidth}
      \includegraphics[width=1.0\linewidth]{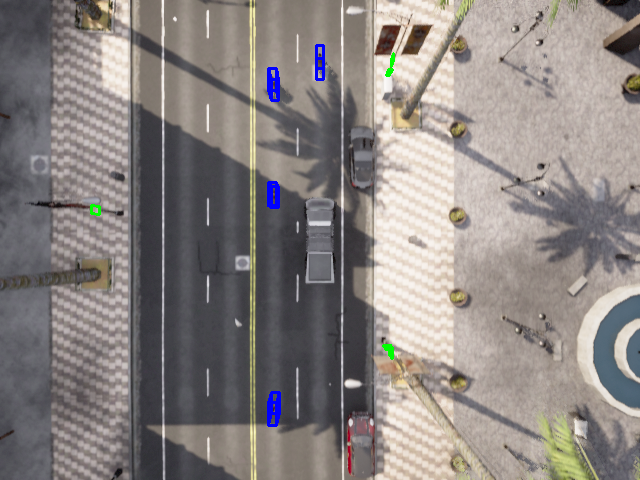}
      \centering $\gls{s:frame}_{9262, 9562}$
    \end{minipage}\newline\vspace{0.5em}
    \begin{minipage}[c]{1.0\linewidth}
      \includegraphics[width=1.0\linewidth]{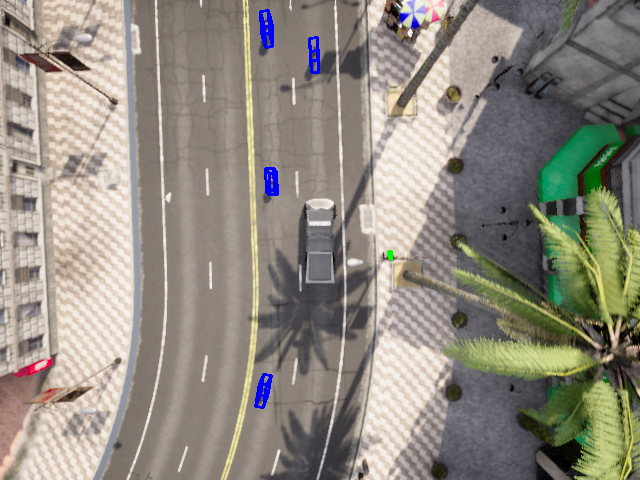}
      \centering $\gls{s:frame}_{12853, 13153}$
    \end{minipage}\vspace{0.5em}
    \caption{A collection of various instances in time within the \glsfmttext{a:carla} simulator where the last frame of match from \cref{query:examples:online:b:1:3} is shown.}
    \label{fig:examples:online:3}
\end{figure}

\added{A illustrative example of matching frames reported by \gls{a:strem} during the simulation are showcased in \cref{fig:examples:online:3}.}

\subsection{Benchmarks}
\label{sec:examples:benchmarks}

\begin{figure*}
    \centering
    \begin{tikzpicture}
        \begin{axis}[
            width=1.0\linewidth,
            height=5.2cm,
            title={Matching Time},
            axis lines=left,
            xlabel={Number of Frames},
            ylabel={Running Time (ms)},
            ymax=300,
            legend pos=north west,
            ymajorgrids,
            grid style=dashed,
        ]
        \addplot [
            color=blue,
            mark=o,
        ]
        coordinates {
            (0,2.082064128077263e-06)(6300,4.071644000167021)(12600,8.371319177777176)(18900,12.50181391688853)(25200,16.645427001964887)(31500,20.76770544978983)(37800,25.053435434839464)(44100,29.155700840822895)(50400,33.258065813625116)(56700,37.468753035494714)(63000,41.60653638246031)(69300,45.68497226877315)(75600,49.74386842198412)(81900,53.79158119059885)(88200,58.26630231909126)(94500,61.85168797893298)(100800,65.83988852238096)(107100,70.04978778301091)(113400,73.98267391858631)(119700,78.34373654982993)(126000,82.4659348480839)(132300,86.46127350421767)(138600,90.64302436107144)(144900,94.88456187886243)(151200,98.97044460314153)
        };
        \addlegendentry{\cref{query:examples:offline:1}}
        \addplot [
            color=red,
            mark=o,
        ]
        coordinates {
            (0,2.122948384377788e-06)(6300,8.607407940862936)(12600,19.104508049169404)(18900,27.932965663498013)(25200,40.86628257975057)(31500,52.203119417788606)(37800,62.283986338315685)(44100,69.80551652539187)(50400,77.91029516742061)(56700,89.33002329615645)(63000,101.29960406142196)(69300,109.92641685581746)(75600,117.63002861501586)(81900,125.69062111500796)(88200,135.51767477761902)(94500,142.6523890076587)(100800,149.689264062877)(107100,161.03012626863097)(113400,166.6565696294246)(119700,174.1911101684028)(126000,182.5429670768254)(132300,192.47029742325395)(138600,205.418548935291)(144900,217.65299247268518)(151200,224.79715342214288)
        };
        \addlegendentry{\cref{query:examples:offline:2}}
        \addplot [
            color=green,
            mark=o,
        ]
        coordinates {
            (0,2.1440463833771113e-06)(6300,6.001555310318332)(12600,12.09181447104296)(18900,18.103847232489755)(25200,24.081892784829197)(31500,29.9730827902736)(37800,35.898117023742564)(44100,41.84852073361961)(50400,47.78646274748677)(56700,53.716433118924954)(63000,59.5178029904881)(69300,65.33811371602734)(75600,72.38157800037202)(81900,78.31190032559525)(88200,84.16656568045917)(94500,89.11599824281178)(100800,94.9727126521627)(107100,101.08215973767194)(113400,106.8127595880291)(119700,113.80051857692065)(126000,119.80452919065871)(132300,124.73912331441271)(138600,130.73724144498414)(144900,136.84111346795638)(151200,142.23423541738092)
        };
        \addlegendentry{\cref{query:examples:online:b:1:1}}
        \addplot [
            color=orange,
            mark=o,
        ]
        coordinates {
            (0,2.0724675697424668e-06)(6300,10.3272545960003)(12600,22.925875515269507)(18900,34.600303569561014)(25200,49.00458133535053)(31500,62.63396580324956)(37800,72.58682144445436)(44100,82.73984195412699)(50400,95.97839857584658)(56700,109.29802777474603)(63000,121.98854882718253)(69300,132.33360506023809)(75600,142.35266384690476)(81900,152.57436392257938)(88200,165.88863626154762)(94500,173.98853447029762)(100800,184.5462176423148)(107100,196.3946114717725)(113400,204.645995381045)(119700,221.65744635149474)(126000,232.6373959346296)(132300,241.89753871247356)(138600,248.00614759972223)(144900,259.50667544541005)(151200,268.13002680677255)
        };
        \addlegendentry{\cref{query:examples:online:b:1:2}}
        \end{axis}
    \end{tikzpicture}
    \caption{Running time performance of \glsfmttext{a:strem}.}
    \label{fig:examples:benchmarks:1}
    \vspace{-10pt}
\end{figure*}
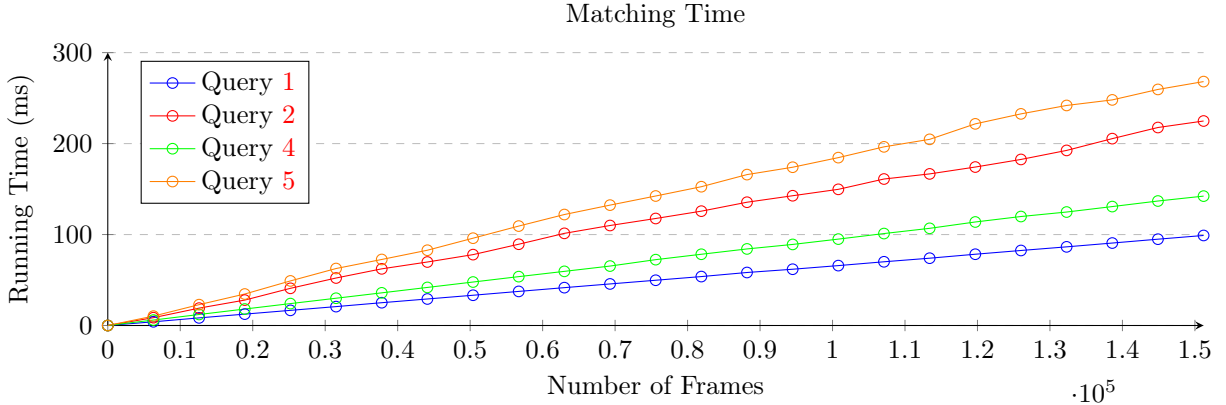

To evaluate the performance of the \gls{a:strem} tool, we ran several different queries against the \gls{k:dataset} dataset.
For each query, the average running time of 10 samples of the matching algorithm was evaluated against 0 to 150K frames.
The results are summarized in \cref{fig:examples:benchmarks:1}.
The benchmarks were ran on a Linux workstation running Fedora 37 (6.2.14-200.fc37.x86\_64) with an AMD Ryzen 7 Pro 4750U processor with Radeon Graphics, 32 GBs of RAM, and a \gls{k:wall-clock} time of 30 seconds.

%% file: src/sections/literature.tex
\section{Related Work}
\label{sec:literature}

The problem of querying video/multimedia datasets has a long history.
Among the earliest works, \cite{delbimbo1995symbolic} presents a spatio-temporal logic that can encode relationships among objects within image sequences.
More recently, the \gls{a:veql} was proposed in \cite{yadav2019vidcep} (the paper also contains an exhaustive review of other video query languages).
\gls{a:veql} is a declarative language similar to SQL and it is used for monitoring of video data streams. 
It supports some ad hoc spatial and topological operators and some basic temporal relations  through the \gls{a:aia} \cite{allen1983maintaining}.
Besides the obvious differences of monitoring \gls{a:aia} (\gls{a:aia} can be encoded in \gls{a:ltl} \cite{rocsu2006allen}) versus \gls{a:re} pattern matching, \gls{a:spre} fully incorporates \gls{s:s4u} and it can foundationally support other modal logics of space (and time).
Beyond queries, \gls{a:tqtl} was proposed in \cite{dokhanchi2018evaluating} to enable basic sanity checks over video feeds of automotive systems using object annotations. 
Furthermore, an online monitoring algorithm for \gls{a:tqtl} was presented in \cite{balakrishnan2021percemon}. 

Typically, perception data streams from automotive applications contain not only image sequences, but also data from a range of other sensing modalities, e.g., radar, lidar, infrared, etc.
In \cite{hekmatnejad2022formalizing}, \gls{a:stpl} was introduced which combines \gls{a:tqtl} \cite{dokhanchi2018evaluating} with an extension of the spatial-temporal logic \gls{s:ptlxs4u} \cite{gabelaia2005combining} to support reasoning over spatial conditions such as intersection and distances between bounding boxes. 
In principle, temporal logics could be used for pattern matching after some modifications to their monitoring algorithms, but in practice, their syntax is not well suited for describing patterns. 
In another line of work, a querying method for sim-to-real applications is presented in \cite{kim2022querying} which uses the \glsentryname{a:scenic} probabilistic programming language \cite{fremont2019scenic}.
Abstract static scenarios of interest are expressed in \glsentryname{a:scenic} which are then queried over labeled datasets through a conversion into a \gls{a:smt} problem.
Even though one can envision that the method in \cite{kim2022querying} can eventually be extended to temporal queries, right now it is restricted to static scenes.


%% file: src/sections/conclusion.tex
\section{Conclusion and Future Work}
\label{sec:conclusion}

In this paper, we proposed \gls{a:spre} as a novel querying language for searching over perception streams using an \gls{s:rexs4u} design.
We demonstrated the application of \glspl{a:spre} in the \gls{k:offline} and \gls{k:online} domain through the development of the \gls{a:strem} tool alongside examples of matching over an \gls{a:av} dataset and the \gls{a:ros} and \gls{a:carla} simulators, respectively.
From this, we are able to find up to 20K+ matches in under 296 ms.
As future work, we plan to include support for existential and universal operators in order to support a richer set of behaviors.
